\documentclass[10pt,twocolumn,letterpaper]{article}

\usepackage{cvpr}
\usepackage{times}
\usepackage{epsfig}
\usepackage{graphicx}
\usepackage{amsmath}
\usepackage{amssymb}
\usepackage{subfigure}

\usepackage[pagebackref=false,breaklinks=true,bookmarks=false]{hyperref}
\graphicspath{{figures/}}

\cvprfinalcopy 

\def\cvprPaperID{2177} 

\ifcvprfinal\fi

\begin{document}

\title{Actionness Estimation Using Hybrid Fully Convolutional Networks}
\author{Limin Wang$^{1,3}$ \quad \quad Yu Qiao$^{1}$ \quad \quad Xiaoou Tang$^{1,2}$ \quad \quad Luc Van Gool$^3$ \\
\small $^{1}$Shenzhen key lab of Comp. Vis. \& Pat. Rec.,  Shenzhen Institutes of Advanced Technology, CAS, China \\
\small $^{2}$Department of Information Engineering, The Chinese University of Hong Kong, Hong Kong \\
\small $^{3}$Computer Vision Lab, ETH Zurich, Switzerland \\
}

\maketitle
\thispagestyle{empty}

\begin{abstract}
Actionness \cite{ChenXXC14} was introduced to quantify the likelihood of containing a generic action instance at a specific location. Accurate and efficient estimation of actionness is important in video analysis and may benefit other relevant tasks such as action recognition and action detection. This paper presents a new deep architecture for actionness estimation, called \emph{hybrid fully convolutional network} (H-FCN), which is composed of appearance FCN (A-FCN) and motion FCN (M-FCN). These two FCNs leverage the strong capacity of deep models to estimate actionness maps from the perspectives of static appearance and dynamic motion, respectively. In addition, the fully convolutional nature of H-FCN allows it to efficiently process videos with arbitrary sizes. Experiments are conducted on the challenging datasets of Stanford40, UCF Sports, and JHMDB to verify the effectiveness of H-FCN on actionness estimation, which demonstrate that our method achieves superior performance to previous ones. Moreover, we apply the estimated actionness maps on action proposal generation and action detection. Our actionness maps advance the current state-of-the-art performance of these tasks substantially.
\end{abstract}


\section{Introduction}
\begin{figure}
  \subfigure[RGB]{
  \begin{minipage}[b]{0.32\linewidth}
    \includegraphics[width=\linewidth]{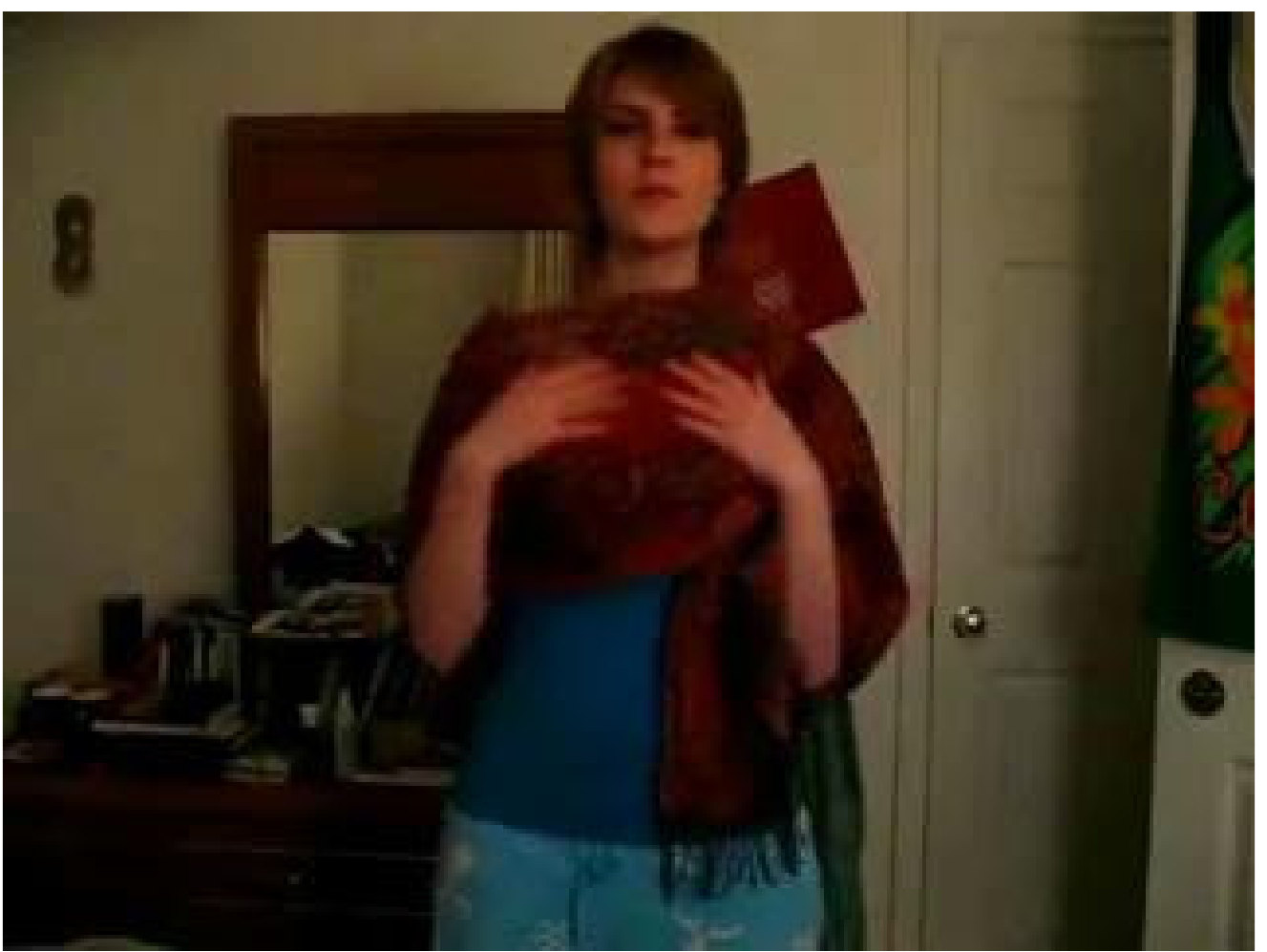}
  \end{minipage}
  }
  \hspace{-3mm}
  \subfigure[Flow-x]{
  \begin{minipage}[b]{0.32\linewidth}
    \includegraphics[width=\linewidth]{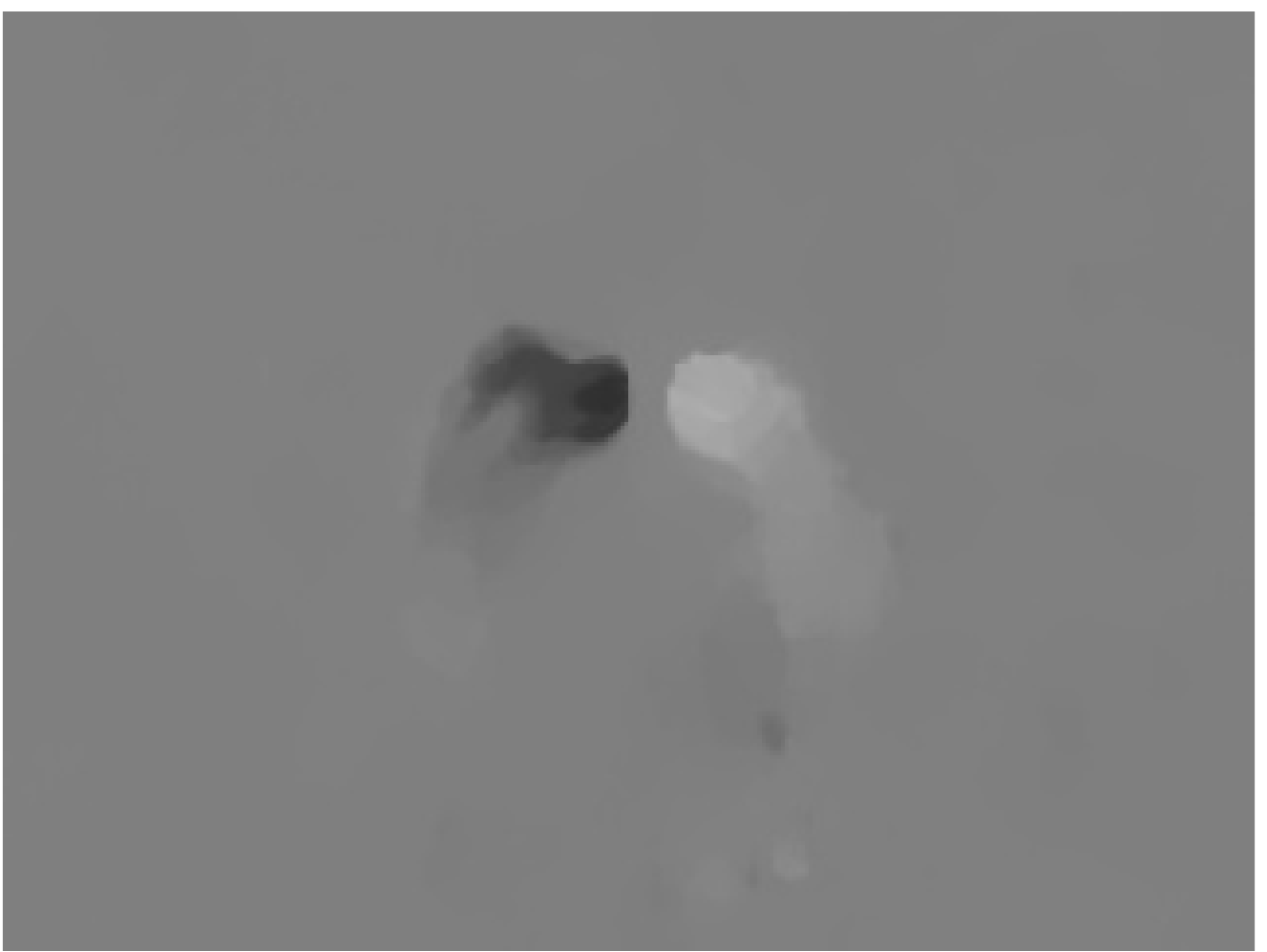}
  \end{minipage}
  }
  \hspace{-3mm}
  \subfigure[Flow-y]{
  \begin{minipage}[b]{0.32\linewidth}
    \includegraphics[width=\linewidth]{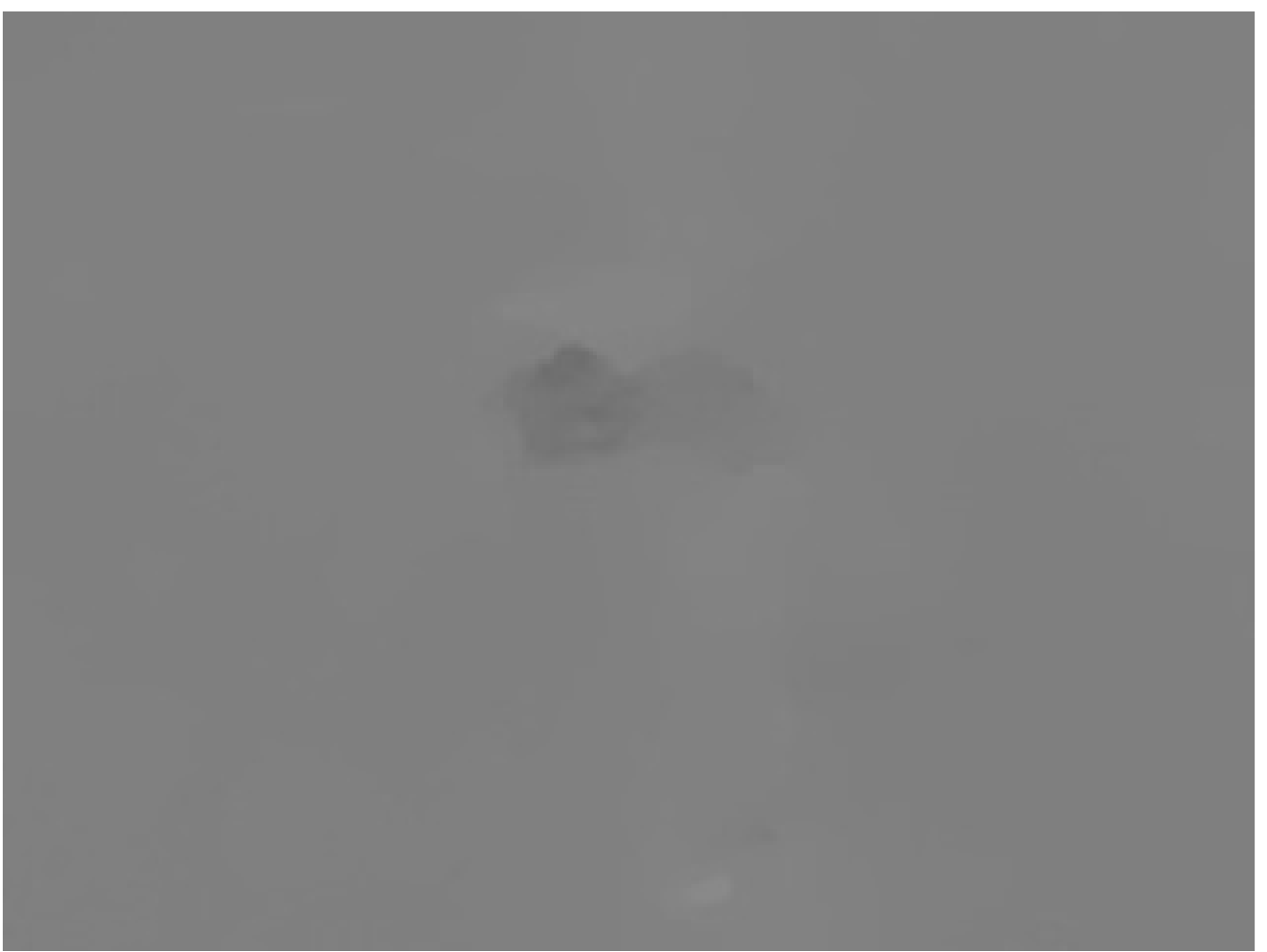}
  \end{minipage}
  }
  \\ \vspace{-1mm}
  \subfigure[A-FCN Result]{
  \begin{minipage}[b]{0.32\linewidth}
    \includegraphics[width=\linewidth]{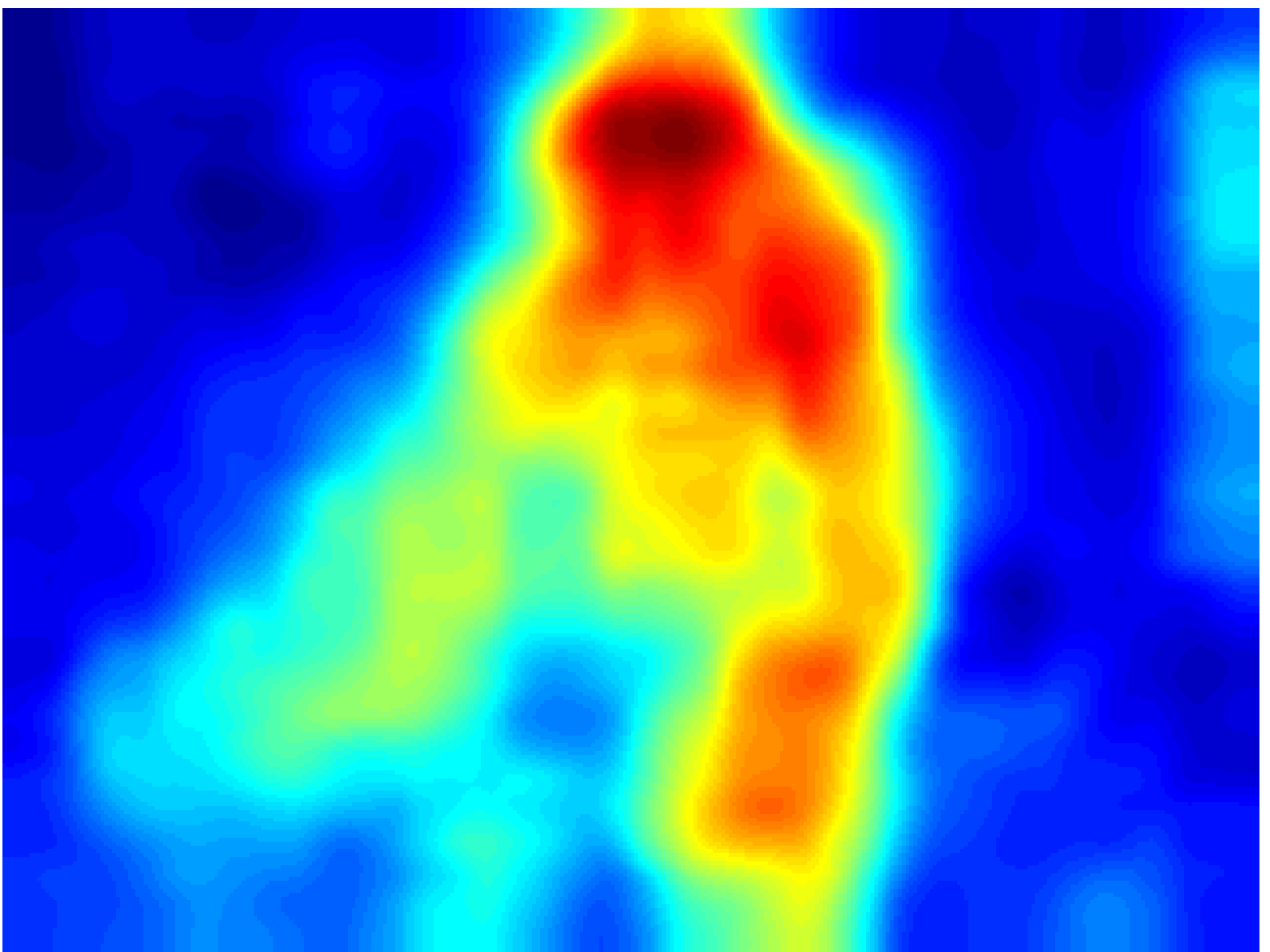}
  \end{minipage}
  }
  \hspace{-3mm}
  \subfigure[M-FCN Result]{
  \begin{minipage}[b]{0.32\linewidth}
    \includegraphics[width=\linewidth]{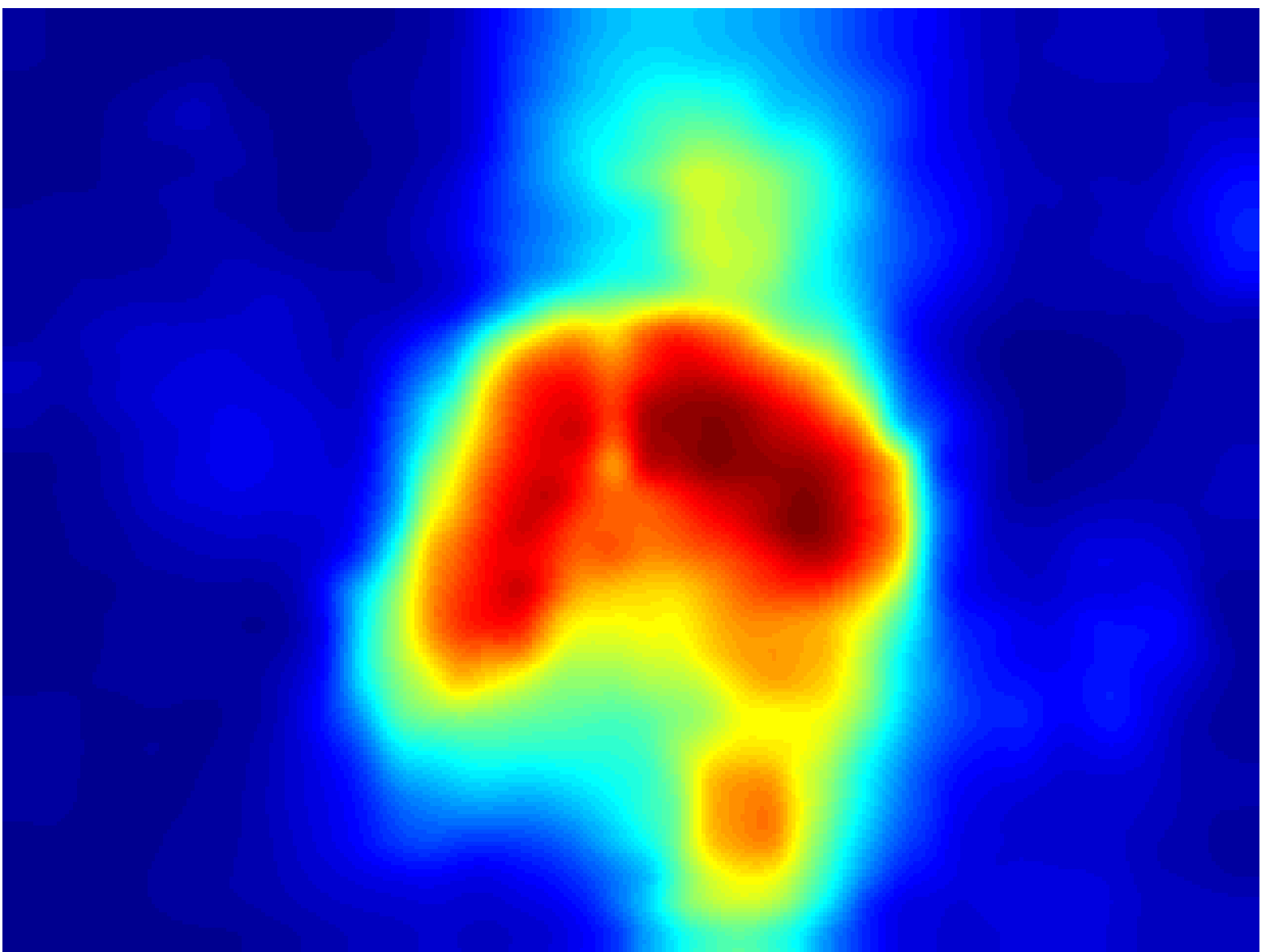}
  \end{minipage}
  }
  \hspace{-3mm}
  \subfigure[H-FCN Result]{
  \begin{minipage}[b]{0.32\linewidth}
    \includegraphics[width=\linewidth]{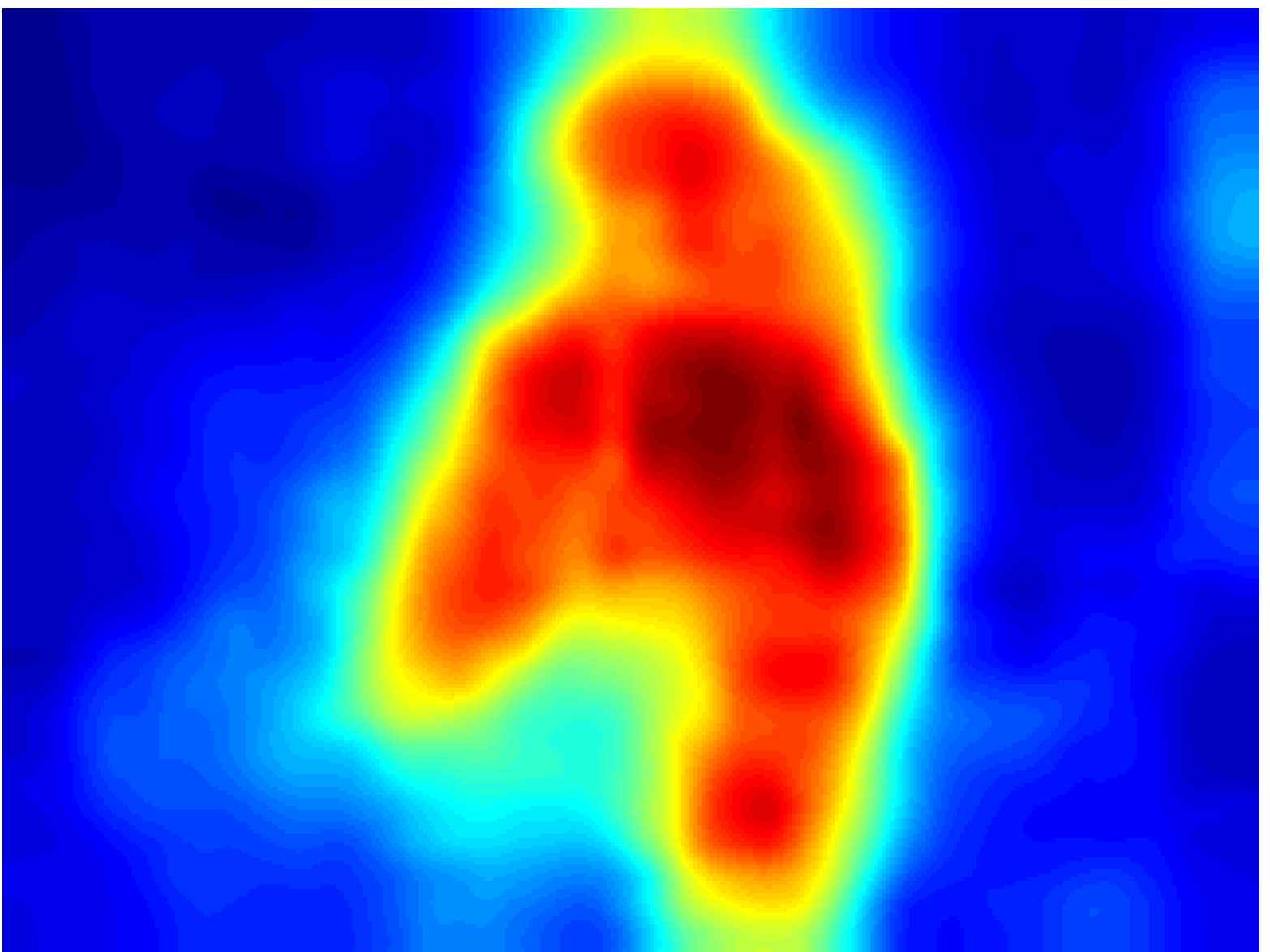}
  \end{minipage}
  }
  \vspace{1mm}
  \caption{\textbf{An example of actionness maps}. Our Hybrid-FCN (H-FCN) is composed of Appearance-FCN (A-FCN) and Motion-FCN (M-FCN). A-FCN captures appearance information from static RGB image, while M-FCN extract motion cues from optical flow fields. The two FCNs are complementary to each other for the task of actionness estimation.}
  \vspace{-3mm}
  \label{fig:illustration}
\end{figure}

Action understanding in videos is an important problem in computer vision and has received extensive research attention in this community rencently. Most of the research works focused on the problem of action classification \cite{Gan2015,Gan16,Laptev05,WangQT13a,WangQT14b,WangQT15b}, which aims at predicting an action label given a video. State-of-the-art classification methods \cite{SimonyanZ14,WangS13a,WangQT15a,WangXW015} have achieved relatively good performance on several challenging datasets, such as HMDB51 \cite{KuehneJGPS11} and UCF101 \cite{Soomro12}. However, these classification methods are only able to answer ``is there an action of certain type present in the video'', but fail to provide the information about ``where is it if there is an action in the video''. To overcome this limitation, the problem of action detection has been studied by several recent works \cite{GkioxariM14,JainGJBS14,TianSS13,WangQT14a}, but these methods still perform relatively poorly on the realistic datasets, such as UCF Sports \cite{RodriguezAS08} and JHMDB \cite{JhuangGZSB13}.

For action detection in videos, we need to estimate bounding boxes of the action of interest at each frame, which together form a spatio-temporal tube in the input video. Sliding window becomes computationally prohibitive due to the huge numbers of candidate windows in the video space. For example, give a video of size $W \times H \times T$, the number of possible boxes for each frame is around $\mathcal{O}((WH)^2)$  and the number of possible tubes for the video is as large as  $\mathcal{O}((WH)^{2T})$. Motivated by fast object detection using proposals \cite{GirshickDDM14}, the idea of ``action proposal'' \cite{OneataRVS14,YuY15} has been introduced for efficient action detection \cite{GkioxariM14,JainGJBS14}. Like object proposal algorithms, most of these methods depend on low-level visual cues, such as spatial edges and motion boundaries, and generate action candidates by hierarchically merging super-voxels \cite{XuC12}. Therefore, these methods usually require heuristic designs and sophisticated merging algorithms, which are difficult to be optimized for action detection and may be sensitive to input noise. Besides, a large amount of candidate regions (around 0.1K-1K) are usually generated by these methods for each frame, which still leads to large computational cost in the subsequent processing.

In this paper we focus on a more general problem regarding action understanding and try to estimate the interestingness maps of generic action given the raw frames, called as \emph{actionness estimation} \cite{ChenXXC14}. Each value of the actionness maps describes the confidence of containing an action instance at this place, where higher value indicates larger probability. According to the recent work \cite{ChenXXC14}, from the perspective of computer vision, action is defined as \textbf{intentional bodily movement of biological agents (such as people, animals)}. Therefore, there are two important visual cues for actionness estimation: \emph{appearance} and \emph{motion}. Appearance information is helpful to locate the biological agents, while motion information contributes to detect bodily movements. In addition, the visual cues of appearance and motion are complementary to each other and fusing them may lead to more accurate actionness estimation.

To accomplish the goal of actionness estimation, we propose a two-stream fully convolutional architecture to transform the raw videos into the map of actionness, called as \emph{hybrid fully convolutional network} (H-FCN). Our H-FCN is composed of two separate neural networks: (i) appearance fully-convolutional network (A-FCN), taking RGB image as input, which captures the spatial and static visual cues, (ii) motion fully-convolutional neural network (M-FCN), using optical flow fields as input, that extracts the temporal and motion information. The actionness maps from these two different FCNs are complementary to each other as shown in Figure \ref{fig:illustration}. Each FCN is essentially a discriminative network trained in an end-to-end and pixel-to-pixel manner. By using fully-convolutional architecture, our H-FCN allows for input with arbitrary size and produces the actionness map of corresponding size.

Specifically, we adopt the contemporary classification networks (ClarifaiNet \cite{ZeilerF14}) into fully-convolutional architecture and transfer the pre-trained model parameters from the large dataset (e.g. ImageNet \cite{DengDSLL009}, UCF101 \cite{Soomro12}) to the task of actionness estimation by fine tuning. We verify the performance of H-FCN for actionness estimation on both images and videos. For image data, there is no motion information available and we only use the A-FCN to produce the actionness map on the dataset of Stanford40 \cite{YaoJKLGF11}. For video data with human movement, we use the H-FCN to estimate the actionness on the datasets of UCF Sports \cite{RodriguezAS08} and JHMDB \cite{JhuangGZSB13}. The experimental results on these two datasets demonstrate that our proposed actionness estimation method outperforms previous methods.

Moreover, actionness map can be viewed as a new kind of feature and could be exploited to assist many video based tasks such as action classification, action detection, and actor tracking. In this paper we incorporate our estimated actionness maps into the successful RCNN-alike \cite{GirshickDDM14} detection framework to perform action detection in videos. We first design a NMS score sampling method to produce action proposals based on actionness maps for each frame. Then, we choose the two-stream convolutional networks \cite{SimonyanZ14} as classifiers to perform action detection. We extensively evaluate the effectiveness of our proposed method on two tasks: action proposal generation on the datasets of Stanford 40 \cite{YaoJKLGF11} and JHMDB \cite{JhuangGZSB13}, and action detection on the dataset of JHMDB \cite{JhuangGZSB13}. 

\section{Related Works}
\label{sec:related}

\begin{figure*}
\centering
  \includegraphics[width=1\textwidth]{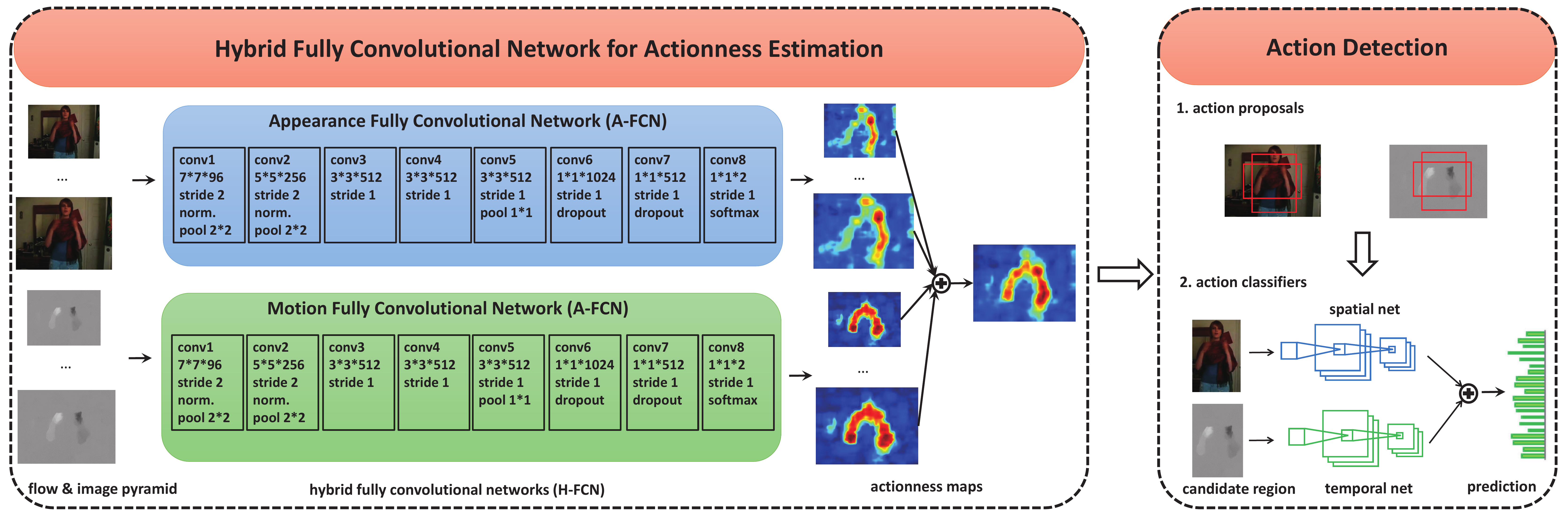}
  \caption{\textbf{Pipeline of our approach}. We propose a new architecture, called hybrid full convolutional network (H-FCN), for the task of actionness estimation. H-FCN contains two parts, namely appearance fully convolutional network (A-FCN) and motion fully convolutional network (F-FCN), which captures the visual cues from the perspectives of static appearance and dynamic motion, respectively. Based the estimated actionness maps, we design a RCNN-alike \cite{GirshickDDM14} action detection system, by first using actionness to generate action proposals and then applying two-stream convolutional networks to classify these proposals.}
  \label{fig:pipeline}
  \vspace{-4mm}
\end{figure*}

\textbf{Actionness and action proposals.} Chen \emph{et al.} \cite{ChenXXC14} first studied the problem of actionness from the philosophical and visual perspective of action. They proposed Lattice Conditional Ordinal Random Fields to rank actionness. Our definition of actionness is consistent with theirs but we introduce a new method called hybrid fully convolutional networks to estimate actionness. Besides, we further apply our actionness map for the task of action detection. Motivated by object proposals in images \cite{AlexeDF12,UijlingsSGS13}, several methods have been developed to generate action proposals in video domain \cite{BerghRBMG13,OneataRVS14,YuY15,JainGJBS14}. Most of these methods generated action proposals based on low-level segmentation and hierarchically merge super-voxels \cite{XuC12} in spatio-temporal domain. However, video segmentation itself is a difficult problem and still under research. Yu \emph{et al.} \cite{YuY15} exploited human and motion detection algorithms to generate candidate bounding boxes as action proposals. Our method does not rely on any pre-processing technique and directly transform raw images into actionness map with fully convolutional networks.

\textbf{Action detection.} Action detection has been comprehensively studied in previous works \cite{LanWM11,KeSH07,YuanLW09,TianSS13,WangQT14a,JainGJBS14,GkioxariM14,LuXC15}. Methods in \cite{YuanLW09,LanWM11} used Bag of Visual Words (BoVWs) representation to describe action and utilized sliding window scheme for detection. Ke \emph{et al.} \cite{KeSH07} utilized global template matching with the volume features for event detection. Lan \emph{et al.} \cite{LanWM11} resorted to latent learning to locate action automatically. Tian \emph{et al.} \cite{TianSS13} extended the 2D deformable part model to 3D cases for localizing actions and Wang \emph{et al.} \cite{WangQT14a} proposed a unified approach to perform action detection and pose estimation by using dynamic-poselets and modeling their relations. Lu \emph{et al.} \cite{LuXC15} proposed a MRF framework for human action segmentation with hierarchical super-voxel consistency. Jain \emph{et al.} \cite{JainGJBS14} produced action proposals using super-voxels and utilized hand-crafted features. Gkioxari \emph{et al.} \cite{GkioxariM14} proposed a similar proposal-based action detection method, but replaced hand-crafted features with deep-learned representations. Our method focuses on actionness estimation and is complementary to these proposal-based action detection methods in sense that our actionness map can be used to generate proposals.

\textbf{Fully convolutional networks.} Convolutional neural networks (CNNs) have achieved remarkable successes for various tasks, such as object recognition \cite{KrizhevskySH12,SimonyanZ14a,SzegedyLJSRAEVR14,ZeilerF14}, event recognition \cite{Xiong15,Gan15}, crowd analysis \cite{shao2015attribute,shao2016slicing} and so on. Recently, several attempts have been made in applying CNN for action recognition from videos \cite{SimonyanZ14,KarpathyTSLSF14,WangQT15a,ZhangWQWW16}. Fully convolutional networks (FCN) were originally introduced for the task of semantic segmentation in images \cite{LongSD14}. One advantage of FCN is that it can take input of arbitrary size and produce semantic maps of corresponding size with efficient inference and learning. To our best knowledge, we are the first to apply FCN into video domain for actionness estimation. In particular, we propose an effective hybrid fully convolutional network which leverages both appearance and motion cues for detecting actionness.


\section{Our Approach}
\label{sec:method}

In this section, we introduce our approach for actionness estimation and show how to apply actionness maps for action proposal generation and action detection. In particular, a brief introduction of fully convolutional networks is firstly described as preparation. Then, we propose hybrid fully convolutional networks to estimate the actionness map from raw frames and optical flow fields. Finally, based on the estimated actionness maps, we develop a RCNN-alike \cite{GirshickDDM14} framework for action detection in videos.

\subsection{Fully convolutional networks}

The feature map processed in each convolutional layer of CNN can be seen as a three-dimensional volume of size $h \times w\times c$, where $h$ and $w$ are the height and width of the map respectively, and $c$ is the number of map channels (filters). The input of CNN is a raw image, with $h\times w$ pixels and $c$ colors. The basic components in CNN contain \emph{convolutional operation}, \emph{pooling operation}, and \emph{activation function}. These basic operations are performed at specific local regions and their parameters are shared across the whole spatial domain of input image or feature map. Hence, this structure allows CNN to have the desired property of translation invariance.

Let $\mathbf{f}_{i,j}^t \in \mathbb{R}^{c_t}$ denote the feature vector at location $(i,j)$ in a particular layer $t$, and $\mathbf{f}_{i,j}^{t+1}$ be the feature vector of following layer $t+1$ at location $(i,j)$. Then, we obtain the following formula for the basic calculation:
\begin{equation}
  \mathbf{f}_{i,j}^{t+1} = h_{k,s}(\{\mathbf{f}^t_{si+\Delta i,sj+\Delta j}\}_{0 \leq \Delta i, \Delta j \leq k}),
  \label{equ:filter}
\end{equation}
where $k$ is the kernel size, $s$ is the stride, and $h_{k,s}$ determines the layer type: matrix multiplication for convolutional layer, average or max operation for pooling layer, an element-wise nonlinear operation for activation function. When deep convolutional networks are constructed by stacking these basic components layer by layer, a network that only contains the nonlinear filter in Equation (\ref{equ:filter}) is called \emph{fully convolutional network} (FCN) \cite{LongSD14}. A FCN can be viewed as performing convolutional operation with a deep filter built by a series of local filters, whose receptive field is determined by the network connections.

We can convert these successful classification convolutional architectures, for example AlexNet \cite{KrizhevskySH12}, ClarifaiNet \cite{ZeilerF14}, GoogLeNet \cite{SzegedyLJSRAEVR14}, and VGGNet \cite{SimonyanZ14a}, into fully convolutional networks by replacing the top fully connected layers with convolutional layers. This replacement leads to two advantages: (i) It allows for input of arbitrary sizes and outputs the corresponding-sized semantic map. (ii) It is very efficient for processing images of large sizes compared with applying sliding window with these classical CNNs. Thanks to these advantages, we choose the architecture of fully convolutional networks for actionness estimation, with a loss function defined as follows:
\begin{equation}
\vspace{-2mm}
\ell(\mathbf{x},\mathbf{m};\theta) = \sum_{i,j} \ell'(\mathbf{x},\mathbf{m}_{ij};\theta),
\label{equ:loss}
\end{equation}
where $\mathbf{x}$ is the input, $\mathbf{m}$ is dense map we need to estimate, and $\theta$ is model parameter. The loss is a sum of each individual loss $\ell'(\mathbf{x},\mathbf{m}_{ij};\theta)$ at a specific location $(i,j)$ over the spatial domain. 

\subsection{Actionness estimation}
\label{sec:h-fcn}

\begin{figure*}
\centering
  \subfigure[Frame image]{
  \begin{minipage}[b]{0.195\linewidth}
    \includegraphics[width=\linewidth]{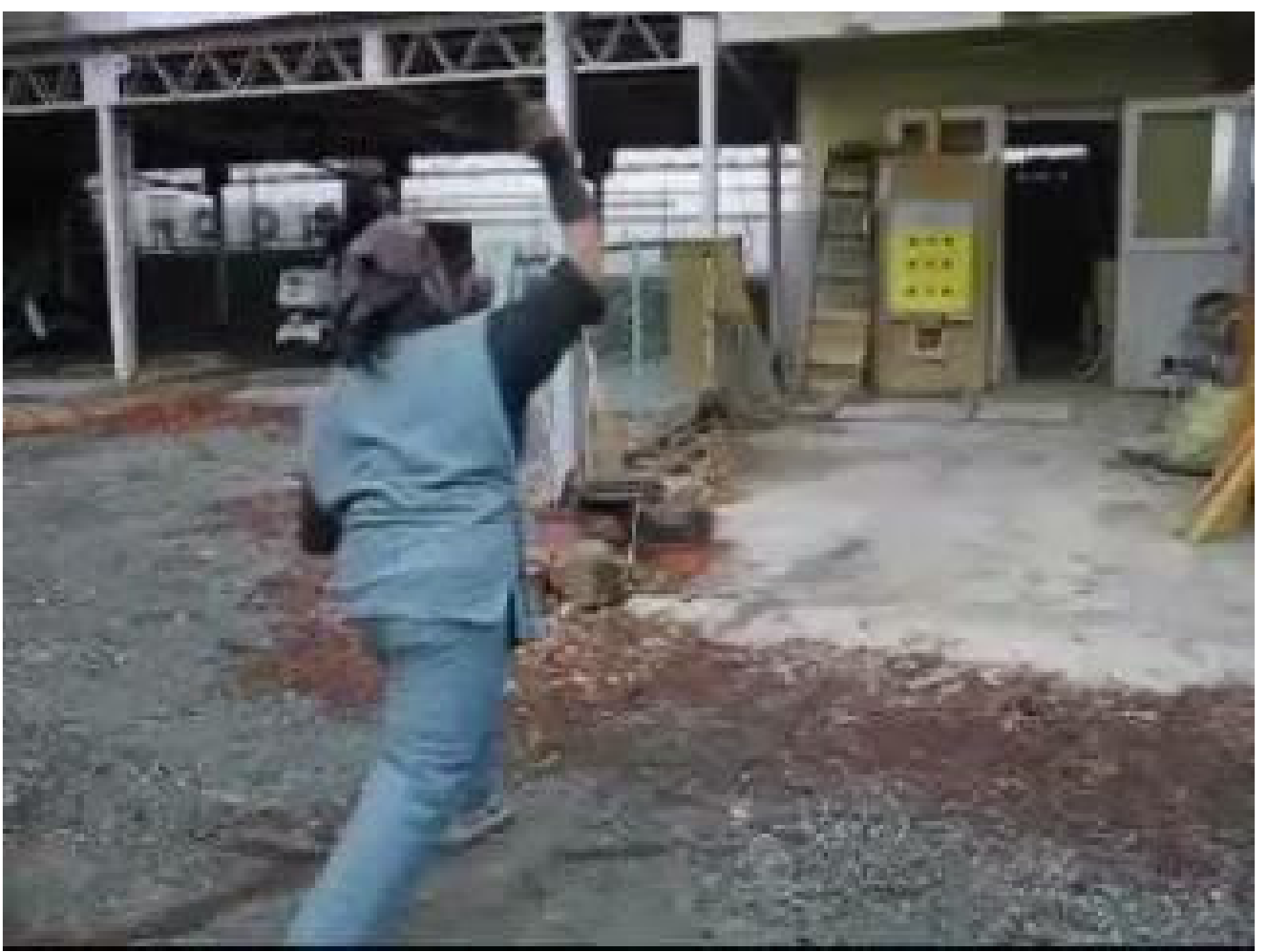}
  \end{minipage}
  }
  \hspace{-3mm}
  \subfigure[Actionness map]{
  \begin{minipage}[b]{0.195\linewidth}
    \includegraphics[width=\linewidth]{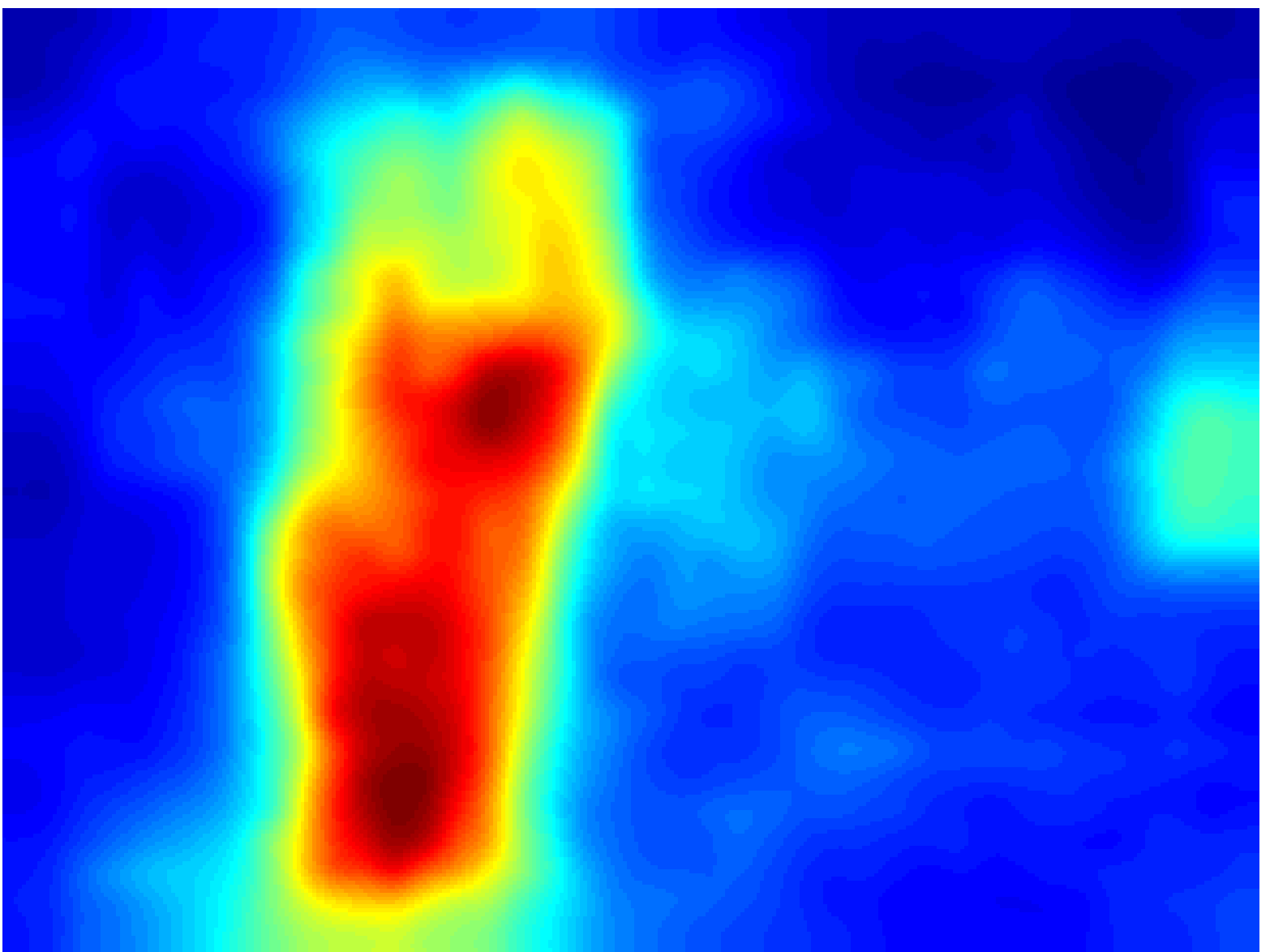}
  \end{minipage}
  }
  \hspace{-3mm}
  \subfigure[Actionness integral image]{
  \begin{minipage}[b]{0.195\linewidth}
    \includegraphics[width=\linewidth]{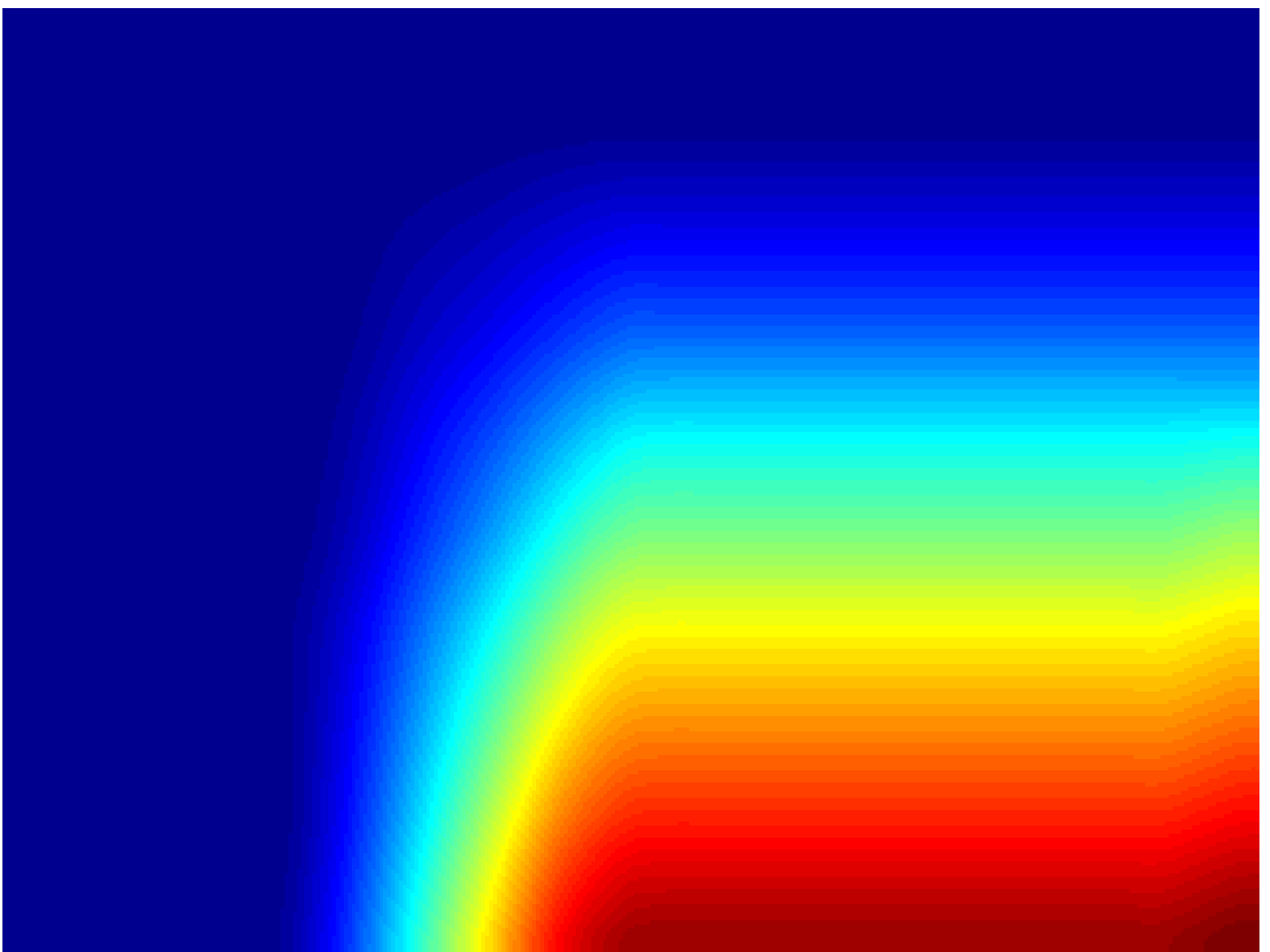}
  \end{minipage}
  }
    \hspace{-3mm}
  \subfigure[Bounding box score]{
  \begin{minipage}[b]{0.195\linewidth}
    \includegraphics[width=\linewidth]{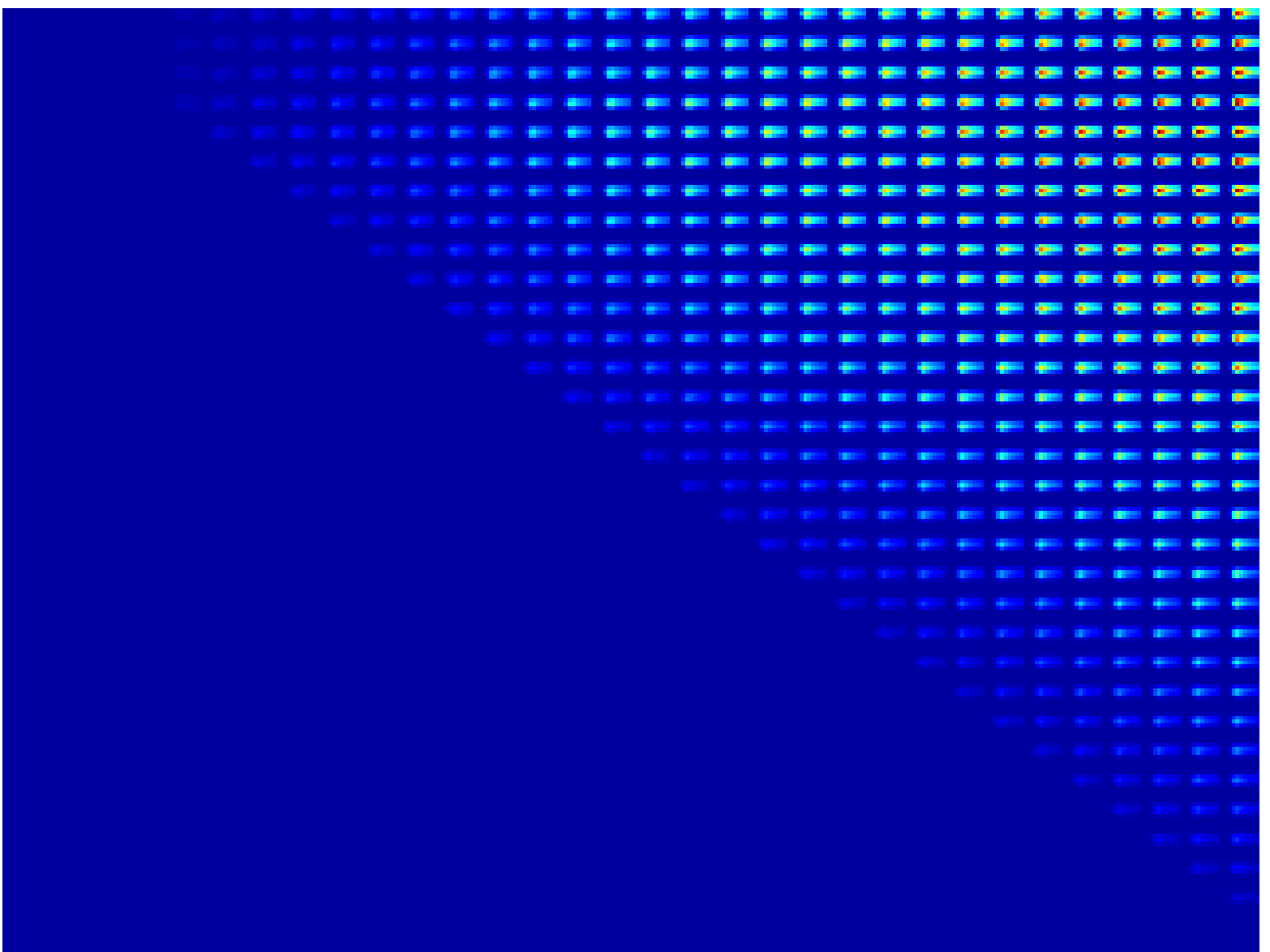}
  \end{minipage}
  }
  \hspace{-3mm}
  \subfigure[Action proposals]{
  \begin{minipage}[b]{0.195\linewidth}
    \includegraphics[width=\linewidth]{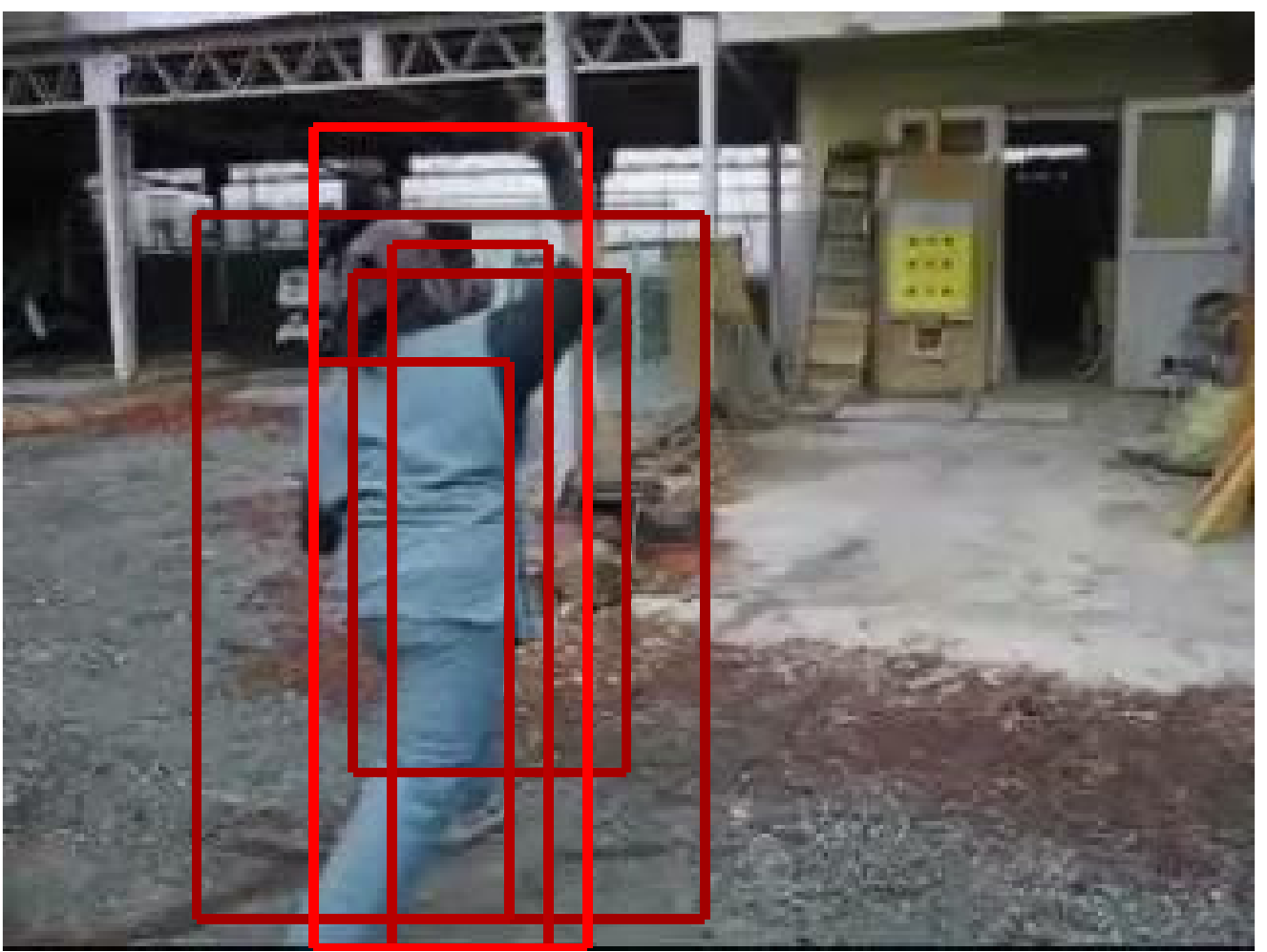}
  \end{minipage}
  }
  \caption{\textbf{Procedure of generating action proposals}. We design an NMS sampling method to generate action proposals based on actionness maps. We resize each map into $32 \times 32$ and compute actionness score of any bounding boxes using integral image representation. Totally, there are $32^4/2$ possible boxes and sample proposal boxes according to their scores and the overlaps between them.}
  \vspace{-4mm}
  \label{fig:actionproposal_pipeline}
\end{figure*}

After the introduction of fully convolutional networks (FCNs), we now describe how to exploit this architecture for the problem of actionness estimation. Actionness essentially describes the likelihood of having an action instance at a certain location. The sizes of action instance vary for different videos and there may be multiple action instances in a single input video. Therefore, it is reasonable to treat the detection of actionness as a dense estimation problem, where the value at each location represents the confidence of containing an action instance there.

Action is defined as intentional bodily movement of biological agents. This definition contains two key elements: (i) ``movement'' and (ii) ``agent''. Bodily movement addresses motion procedure contained in action, while agent refers to the actor performing the action. According to this definition, two visual cues are crucial for estimating actionness, namely \emph{appearance} and \emph{motion}. The motion cues describe the visual patterns of bodily movement and the appearance cues capture the static image information about actors. Following the two-stream convolutional networks \cite{SimonyanZ14} for action recognition, we propose a \emph{hybrid fully convolutional networks} (H-FCN) for the task of actionness estimation, as illustrated in Figure \ref{fig:pipeline}. H-FCN is composed of two networks: \emph{Appearance fully convolutional network} (A-FCN) and \emph{Motion fully convolutional network} (M-FCN).

The appearance fully convolutional network (A-FCN) uses a single frame as input, which is a $W \times H \times 3$ volume. A-FCN aims to learn useful features from the appearance cues for actionness estimation. The input of motion fully convolutional network (M-FCN) is a stack of the optical flow fields from two consecutive frames, thus its size is $W \times H \times 4$. The goal of M-FCN is to extract effective representation from motion patterns. Hopefully, A-FCN and M-FCN capture visual cues from two different perspectives and the combination of them is expected to be more powerful due to their complementarity. The final estimated actionness map is an average of the two maps from A-FCN and M-FCN. Most of action datasets provide only the bounding boxes of action instances instead of the actor segmentation masks. The bounding boxes can be viewed as a kind of weak supervision for actionness map. We convert the annotations of bounding boxes into binary actionness maps, simply by setting the actionness value of pixels inside the bounding box as $1$ and otherwise as $0$. Although this weak supervision is not accurate, we observe that it is sufficient to perform H-FCN training from the experimental results in Section \ref{sec:experiment}.

Specifically, the architectures of A-FCN and M-FCN are similar to each other except for the input layers, and the network details are shown in Figure \ref{fig:pipeline}. Basically, we adapt the successful ClarifaiNet \cite{ZeilerF14} to build our H-FCN. But we make three important modifications to make the network structure more suitable for the task of actionness estimation. {\em First}, we replace the fully connected layers (fc6, fc7, and fc8) with the convolutional layers (conv6, conv7, and conv8), where the kernel size is $1 \times 1$ and convolutional stride is $1 \times 1$. {\em Second}, we change the pooling stride from $2 \times 2$ to $1 \times 1$ after the $5^{th}$ convolutional layer. As our goal is to estimate the dense map of actionness, we need to reduce the down sampling ratio caused by the pooling operation. {\em Third}, the H-FCN output at each position is two-dimensional, since we only need estimate the presence likelihood of an action instance. We choose cross-entropy loss in Equation (\ref{equ:loss}) to train our H-FCN and the implementation details about network training can be found in Section \ref{sec:experiment}.

\textbf{Extension to multi-scale actionness estimation.} The above description on actionness estimation is based on a single scale. However, in realist videos, action instances may vary in scales and we propose an effective and efficient method to handle the issue of scale variance. The fully convolutional nature of H-FCN allows for handling input images of different sizes and producing the actionness maps of corresponding sizes. As shown in Figure \ref{fig:pipeline}, we construct multi-scale pyramid representations of video frames and optical flow fields. We then feed these pyramid representations into H-FCN to obtain multi-scale actionness maps. Finally, these multi-scale actionness maps are resized to the same size and averaged to produce the final estimated maps. In practice, we use $4$ scales for pyramid representations with scale set to $1/\sqrt{2}, 1, \sqrt{2}, 2$. It is worth noting that we just adopt this multi-scale actionness estimation during test phase of H-FCN and we only train H-FCN from a single scale determined by the ground truth.

\subsection{Application on action detection}
\label{sec:detection}

In this subsection we will show how to use the estimated actionness maps for action detection in videos. Generally speaking, our estimated actionness maps can be viewed as new kind of features and can also benefit other relevant problems, such as action classification, actor tracking and so on. More specifically, we adopt an RCNN-alike \cite{GkioxariM14} action detection framework to verify the effectiveness of actionness maps. RCNN-alike action detection framework consists of two steps: generating action proposals and classifying the obtained proposals. Here we aim to design a more unified framework for action detection, where we produce action proposals based the outputs of our H-FCNs rather than using traditional proposal generation method like selective search \cite{UijlingsSGS13}.

\textbf{Action proposals.} Based on actionness maps, we design a simple yet effective method to generate action proposals for each frame. Specifically, in our current implementation, we propose a non-maximum suppression (NMS) sampling method to produce boxes based on actionness map. As shown in Figure \ref{fig:actionproposal_pipeline}, we first resize the actionness map into scale $32 \times 32 $. Then, we use integral image representation to speed up the calculation of average actionness score in the bounding boxes of any sizes. Finally, we sample boxes according to their scores and the spatial overlaps between them. This NMS sampling method has two benefits: sampling boxes with high actionness scores and covering diverse locations.

\textbf{Action classifiers.} Regarding action classifiers, we choose the two-stream convolutional networks \cite{SimonyanZ14} and adapt the pre-trained models to the specific classification task for the target dataset. For positive examples, we crop the frame regions or optical flow fields using the ground truth bounding boxes. For negative examples, we choose these action proposals that overlap less than $0.25$ with ground truth regions. The last layer of two-stream convolutional networks has $|A|+1$ outputs, classifying the action proposals into a pre-defined action category or a background class. At test time, we directly use the $k^{th}$ output of two-stream convolutional networks as the score of $k^{th}$ action detector.


\section{Experiments}
\label{sec:experiment}

In this section, we first introduce the evaluation datasets and their experimental settings. Then, we describe the implementation details of training H-FCNs. Finally, we evaluate our proposed method and perform comparison with other approaches on three tasks, namely actionness estimation, action proposal generation, and action detection.

\subsection{Datasets}

In order to evaluate our proposed method, we conduct experiments on both images and videos. Specifically, we choose three datasets, namely Stanford40 Actions dataset \cite{YaoJKLGF11}, UCF Sports dataset \cite{RodriguezAS08}, and JHMDB dataset \cite{JhuangGZSB13}. The Stanford40 Action dataset contains $9,532$ images of human performing $40$ diverse daily actions, such as riding-bike, playing guitar, calling, and so on. In each image, a bounding box is provided to annotate the actor. The whole dataset is divided into $4,000$ training images and $5,532$ testing images. We use these bounding boxes in training images to learn our A-FCN and the bounding boxes of testing images to evaluate the performance of trained model.

The UCF Sports dataset \cite{RodriguezAS08} is composed of broadcast videos. It has $150$ video clips and contains $10$ action classes, such as diving, golfing, swinging, and so on. It provides the bounding boxes of actors for all the frames. The whole dataset is split into 103 samples for training and $47$ samples for testing. We follow the standard split of training and testing to learn and evaluate our H-FCN.

The JHMDB dataset \cite{RodriguezAS08} is a much larger dataset with full annotations of human joints and body masks, containing $928$ videos and $21$ action classes. The dataset provides three different splits of training and testing, and we report the average performance over these three splits. It should be noted that, like other datasets, we simply use the bounding boxes generated from the body masks as weak supervision to train and evaluate our H-FCN.

The UCF Sports and JHMDB are two large public datasets with bounding box annotations and actionness ground truth. Although these datasets contain temporally trimmed videos, they exhibit complex background and large intra-class variations. Therefore, estimating actionness on these realistic videos are still very challenging.

\subsection{Implementation details}

In this subsection, we describe the training details of the H-FCN introduced in Section \ref{sec:h-fcn} and the two-stream action classifiers introduced in Section \ref{sec:detection}. Training deep convolutional networks is extremely challenging for these action datasets, as the their sizes are much smaller compared with that of the ImageNet dataset \cite{DengDSLL009}. Therefore, we choose the strategy of ``supervised pre-training and careful fine-tuning'' to relieve the over-fitting risk caused by small training data.

For appearance fully convolutional network (A-FCN), we choose the model pre-trained on the ImageNet dataset, which is released by paper \cite{ChatfieldSVZ14}. Then, we transfer the model parameters of convolutional layers to A-FCN and fine tune the network weights on the target dataset for actionness estimation. To reduce the risk of over-fitting, we fix the parameters of the first three convolutional layers and set the learning rate of fourth and fifth convolutional layer as $0.1$ times of network learning rate. The learning rate for the whole network is set as $10^{-2}$ initially, decreased to $10^{-3}$ after 1K iterations, and to $10^{-4}$ after 2K iterations, and training is stopped at 3K iterations. The network weights are learned using the mini-batch (set to 100) stochastic gradient descent with momentum (set to 0.9). During training phase, we resize the training images as $224 \times 224$ and their corresponding actionness map as $14 \times 14$. For testing, we use the multi-scale pyramid representations of images to produce the multi-scale actionness maps as described in Section \ref{sec:h-fcn}. These actionness maps from different scales are first up-sampled to that of original image and then averaged.

For motion fully convolutional network (M-FCN), the input is 3D volume of stacking two-frame optical flow fields. We choose the TVL1 optical flow algorithm \cite{ZachPB07} and use OpenCV implementation, due to its balance between accuracy and efficiency. For fast computation, we discretize the values of optical flow fields into integers and set their range as $0$-$255$ just like images. We choose to pre-train the M-FCN on the UCF101 dataset \cite{Soomro12}, which contains $13,320$ videos and $101$ action classes. We first train the ClarifaiNet on UCF101 from scratch for the task of action recognition. As the dataset is relatively small, we use high dropout ratios to improve the generalization capacity of learned model (0.9 for fc6 layer and 0.8 for fc7 layer). The training procedure of ClarifaiNet on the UCF101 dataset is similar to that of two-stream convolutional networks \cite{SimonyanZ14}. After the training on the UCF101 dataset, we transfer the weights of convolutional layers of ClarifaiNet to M-FCN, and fine tune the whole network on the target dataset for actionness estimation. The fine tuning procedure is the same with that of A-FCN.

The architecture of two-stream action classifier in Section \ref{sec:detection} is the same with that of its original version \cite{SimonyanZ14} except the final output layer. Specifically, we follow our previous works on action recognition with deep learning \cite{WangQT15a,WangXW015}, and the training procedure of two-stream action classifier on target dataset is the same with theirs. The training code with multi-GPU extension is publicly available \footnote{\url{https://github.com/yjxiong/caffe}}.

\begin{figure}
  \includegraphics[width=0.50\linewidth]{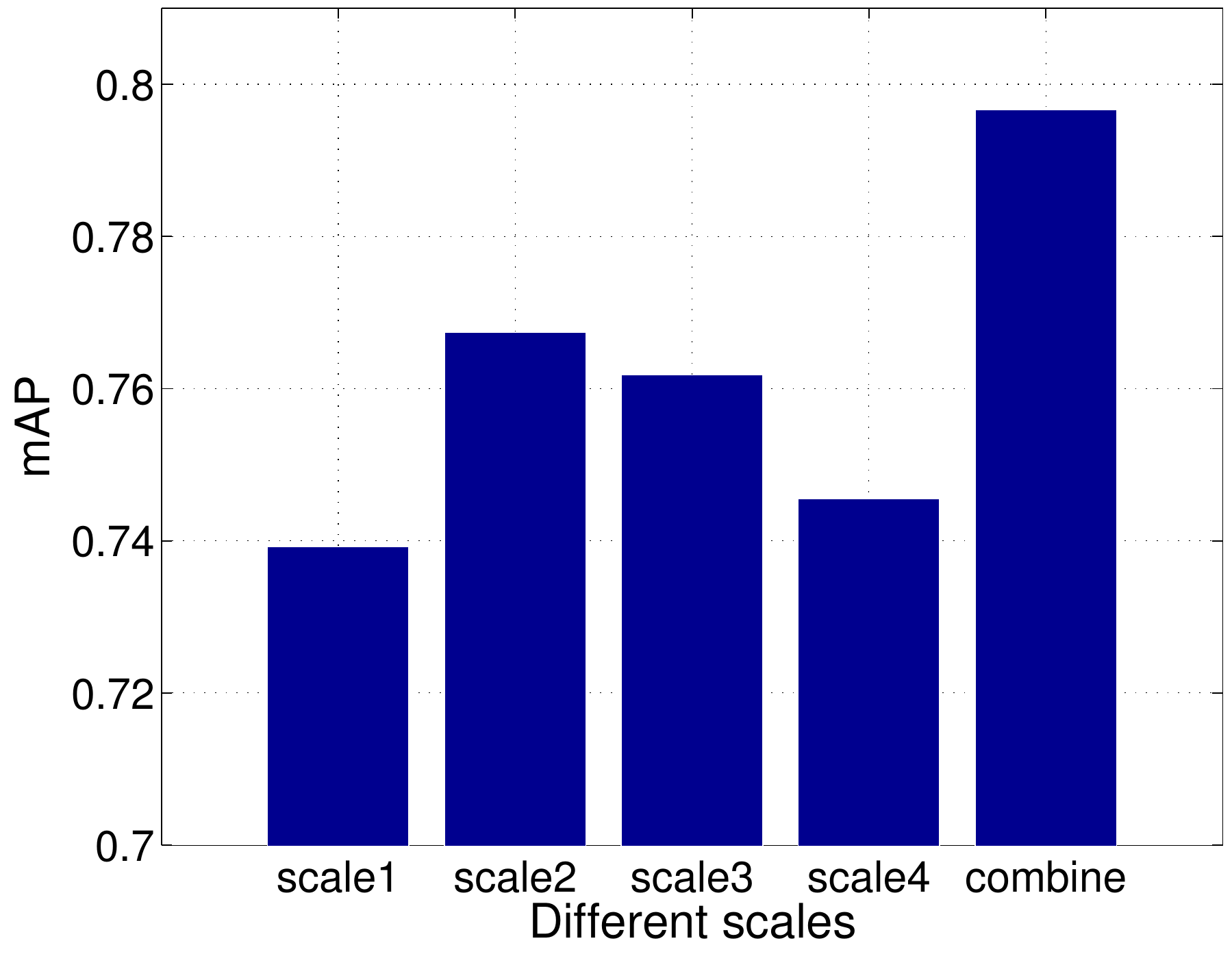}
  \includegraphics[width=0.49\linewidth]{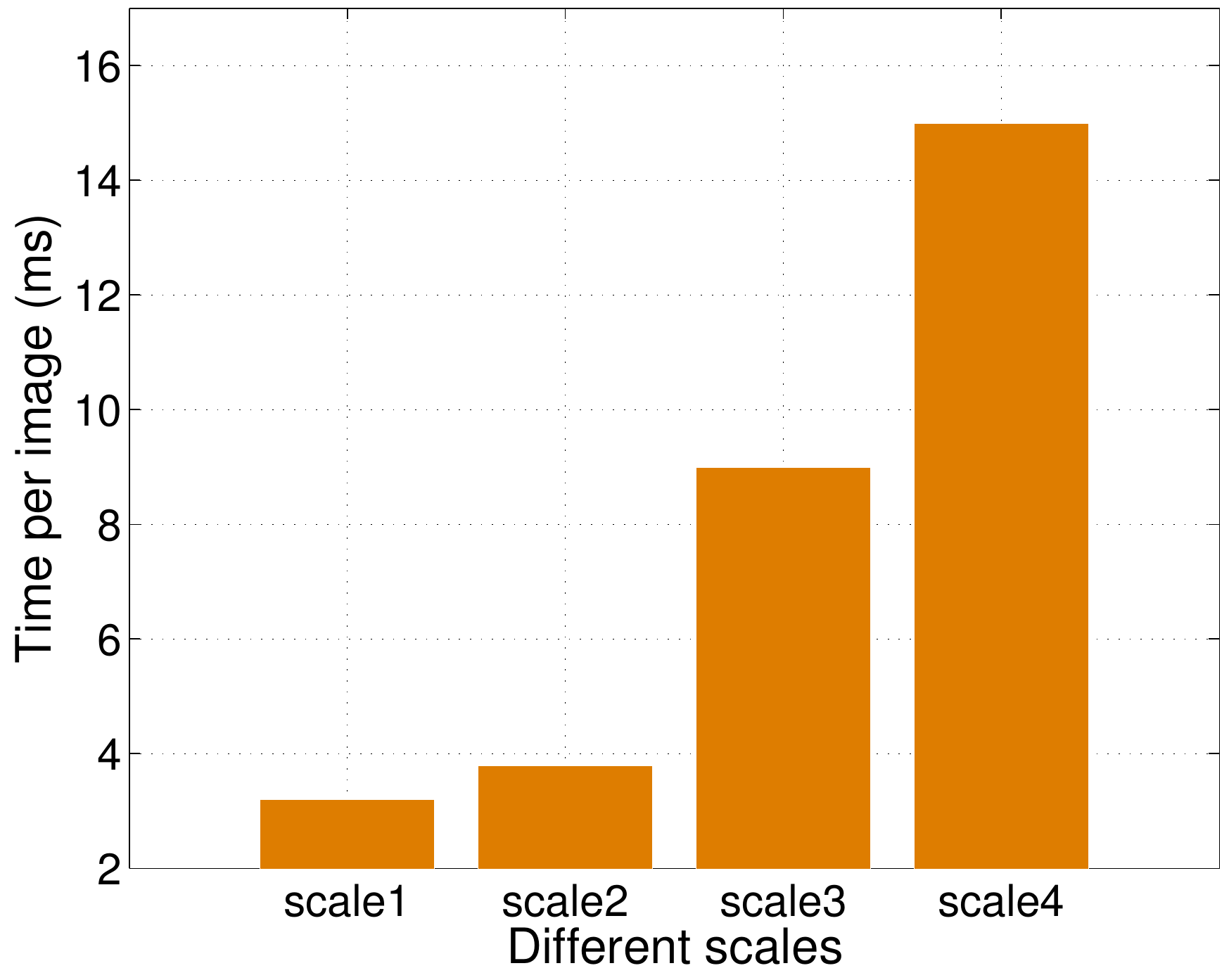}
  \caption{Exploration of multi-scale image representation for actionness estimation on the Standard40 dataset. Left: Performance of different scales and their combination. Right: Computational costs of different scales.}
  \label{fig:multiscale}
  \vspace{-3mm}
\end{figure}

\begin{table}
\small
\centering
  \begin{tabular}{|l|c|c|c|}
    \hline
     Method & Stanford 40 & UCF Sports & JHMDB \\
    \hline
    \hline
    L-CORF \cite{ChenXXC14} & 72.5\% & 60.8\% & 69.1\%\\
    DPM \cite{FelzenszwalbGMR10} & \textbf{85.6\%} & 54.9\% & 58.2\% \\
    RankSVM \cite{joachims1999making} & 55.8\% & 21.9\% & - \\
    MBS \cite{SheikhJK09} & - & 22.8\% & - \\
    \hline
    \hline
    A-FCN & 79.7\% & 75.0\% & 80.7\% \\
    M-FCN & - & 77.2\% & 80.6\% \\
    H-FCN & - & \textbf{82.7\%} & \textbf{86.5\%} \\
    \hline
  \end{tabular}
  \vspace{1mm}
  \caption{Evaluation of actionness estimation. We report mAP values on three datasets and compare with the previous methods.}
  \vspace{-4mm}
  \label{tbl:actionness-evaluation}
\end{table}

\begin{figure*}[t]
\center
  \includegraphics[width=0.121\linewidth]{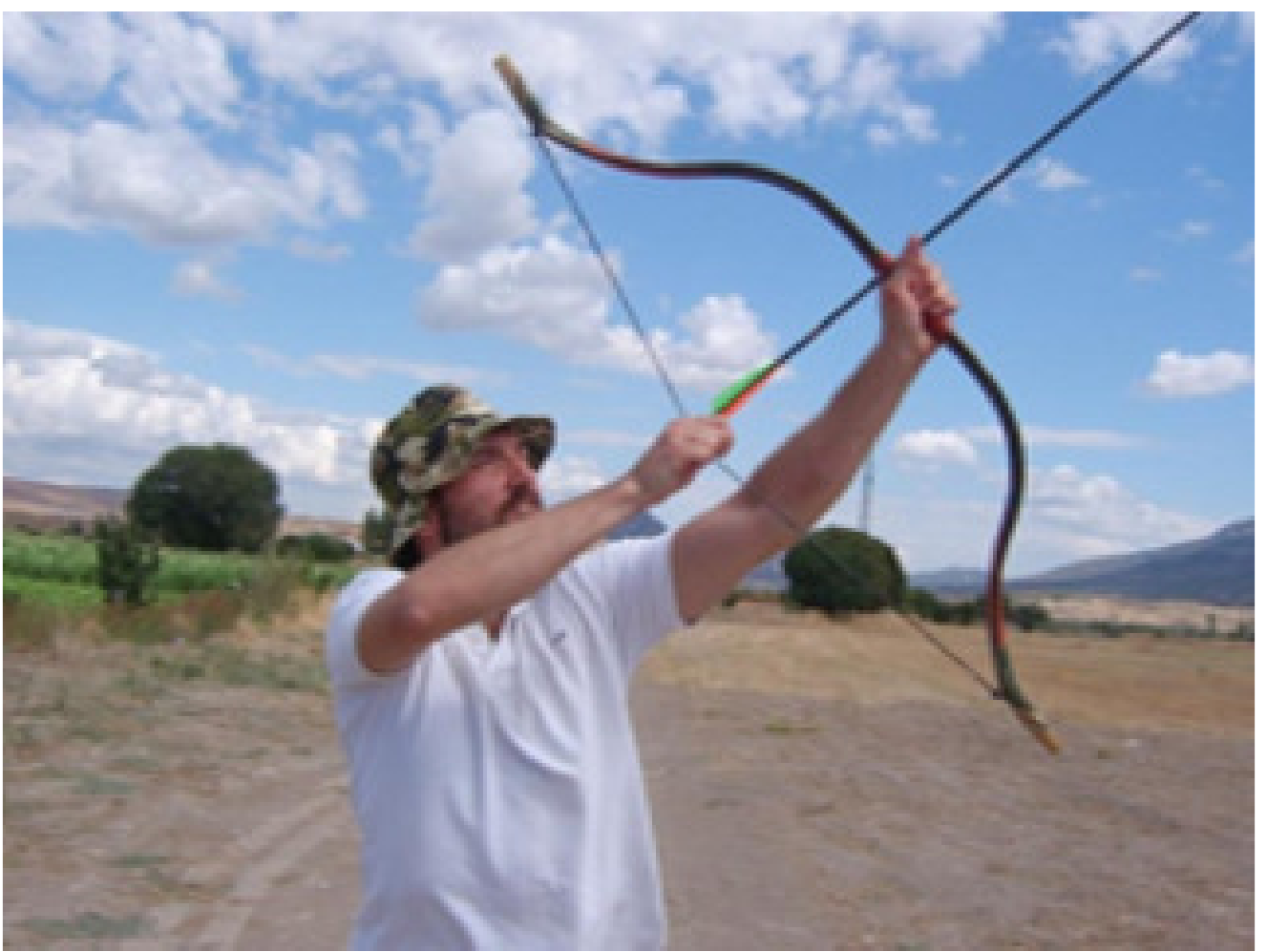}
  \hspace{-1.5mm}
  \includegraphics[width=0.121\linewidth]{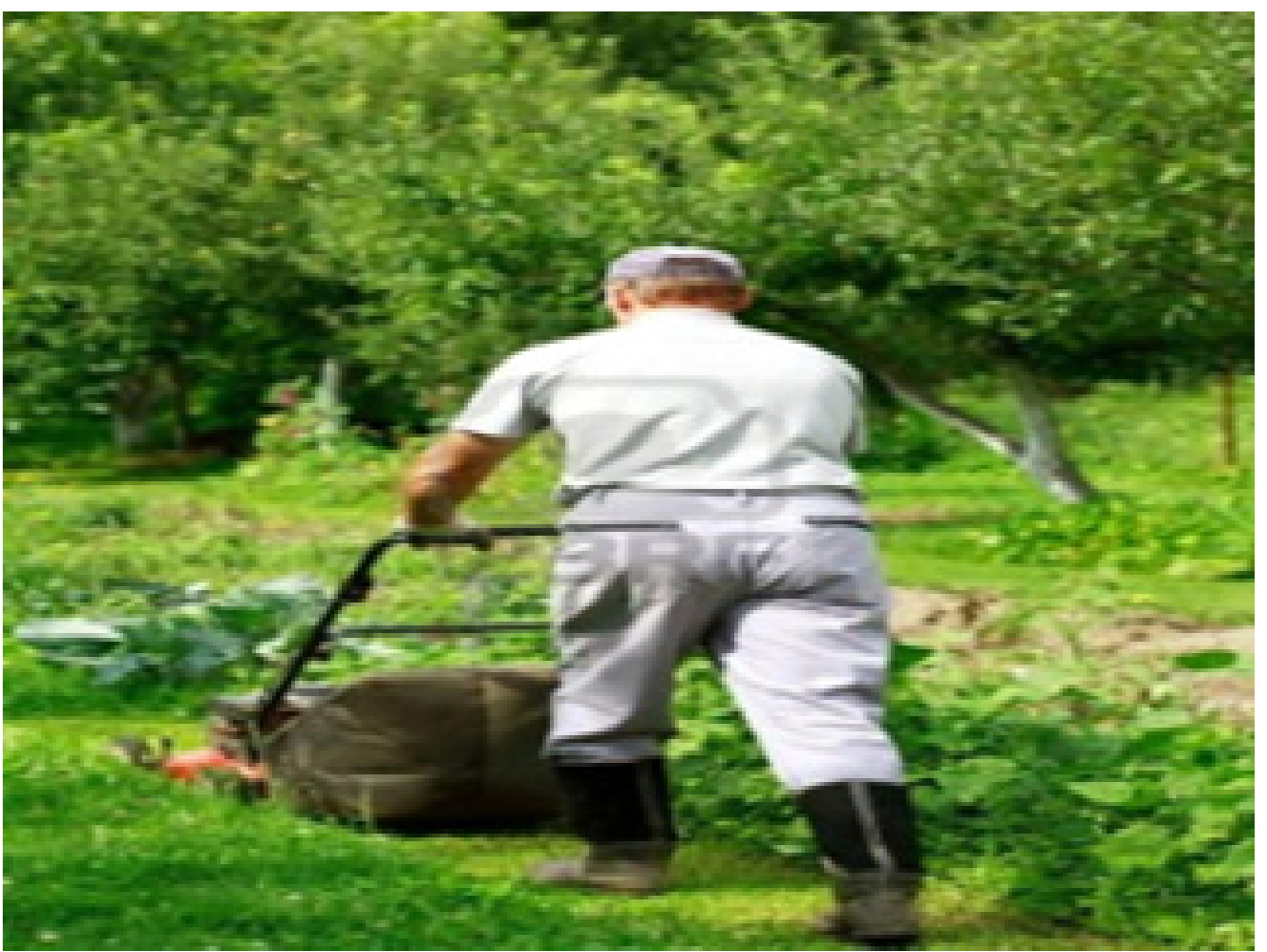}
  \hspace{-1.5mm}
  \includegraphics[width=0.121\linewidth]{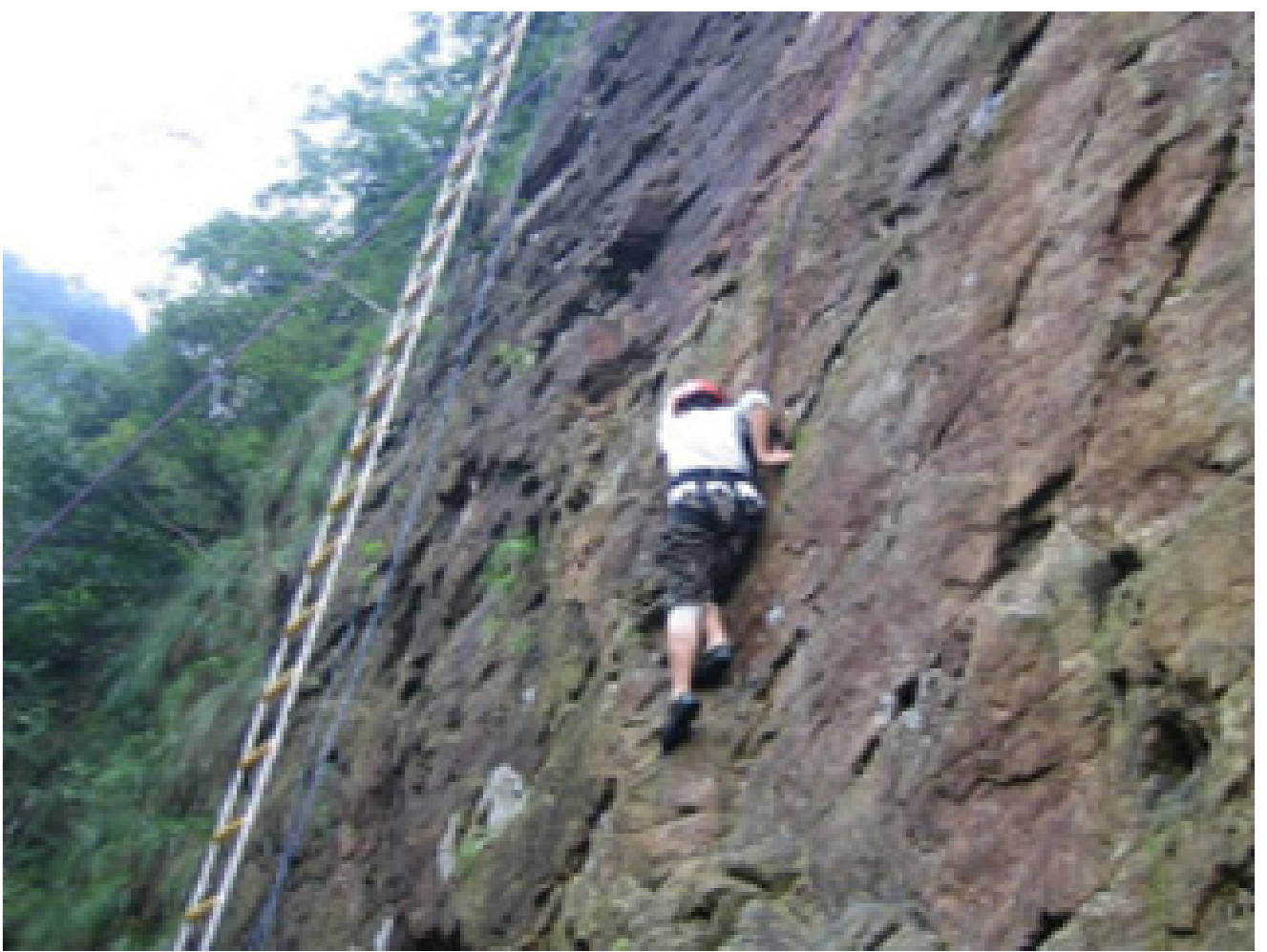}
  \hspace{-1.5mm}
  \includegraphics[width=0.121\linewidth]{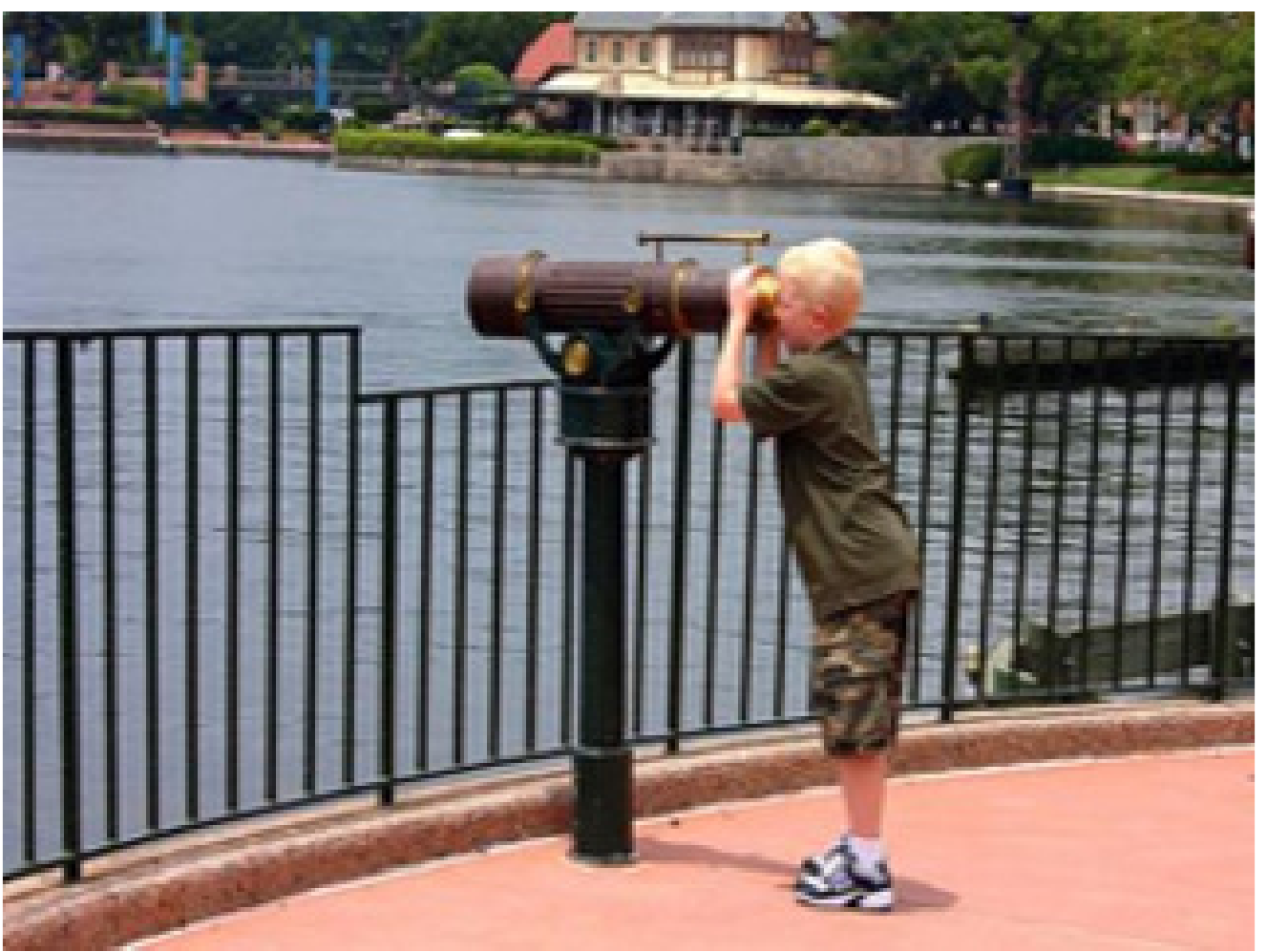}
  \hspace{-1.5mm}
  \includegraphics[width=0.121\linewidth]{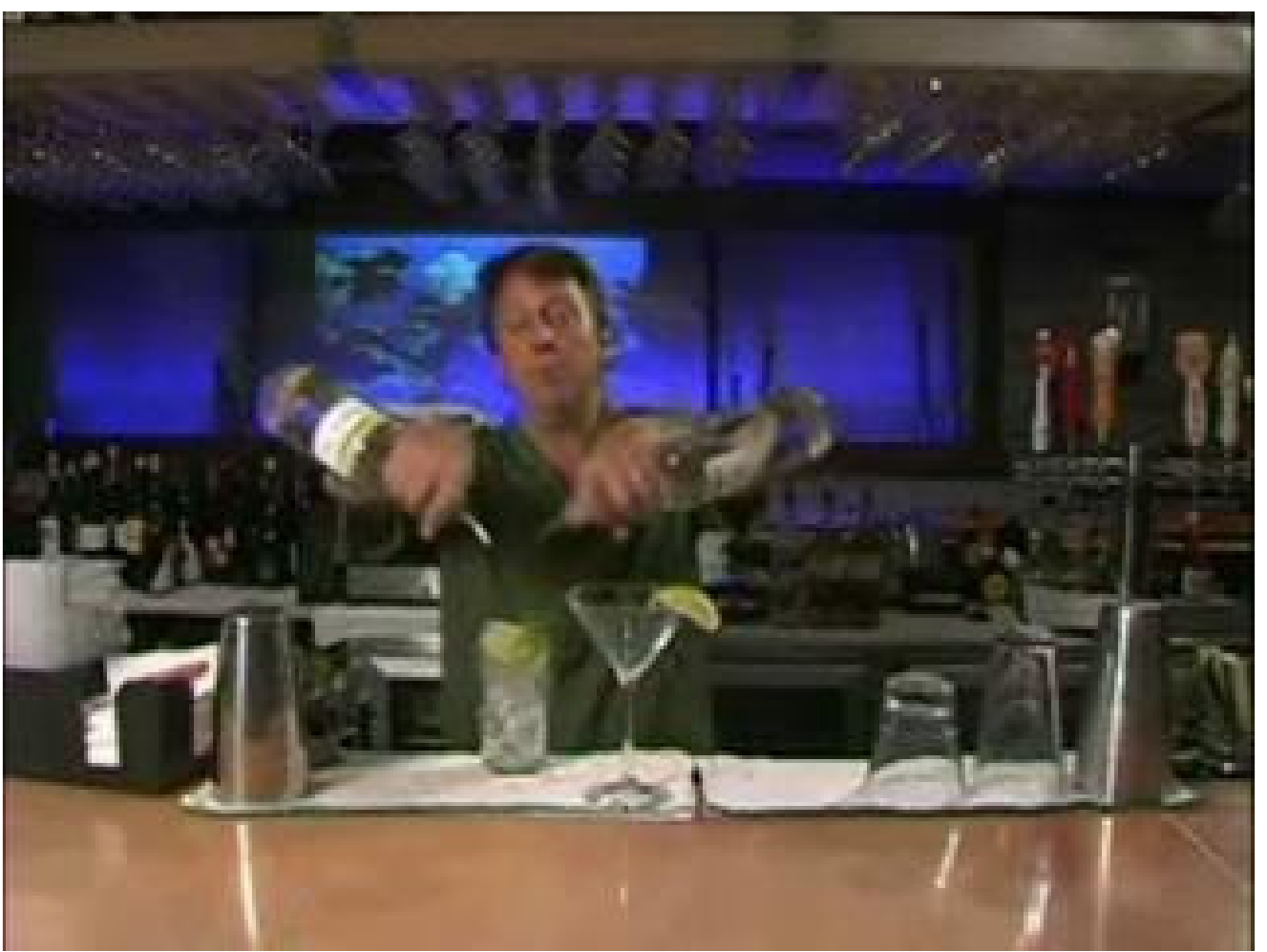}
  \hspace{-1.5mm}
  \includegraphics[width=0.121\linewidth]{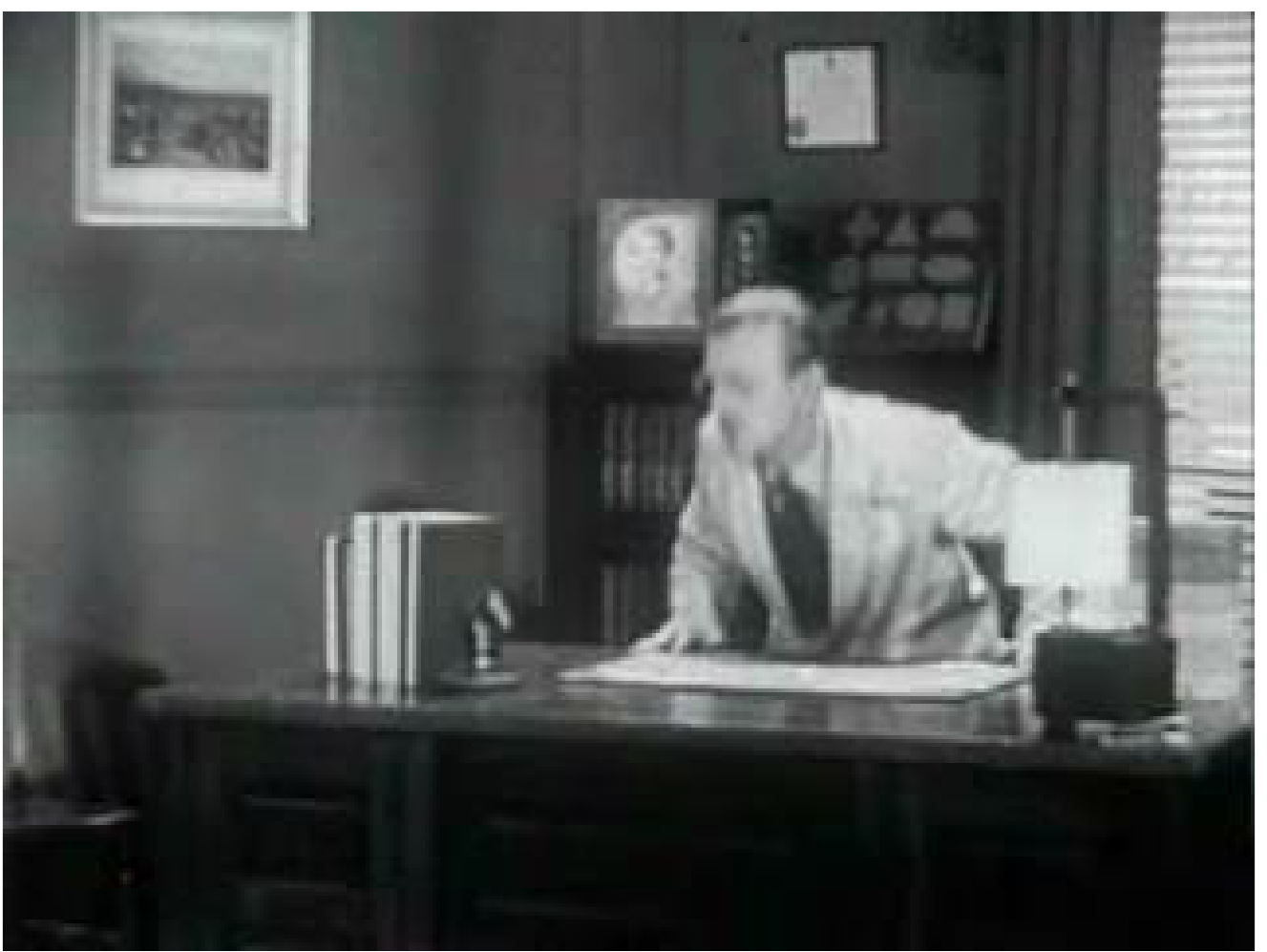}
  \hspace{-1.5mm}
  \includegraphics[width=0.121\linewidth]{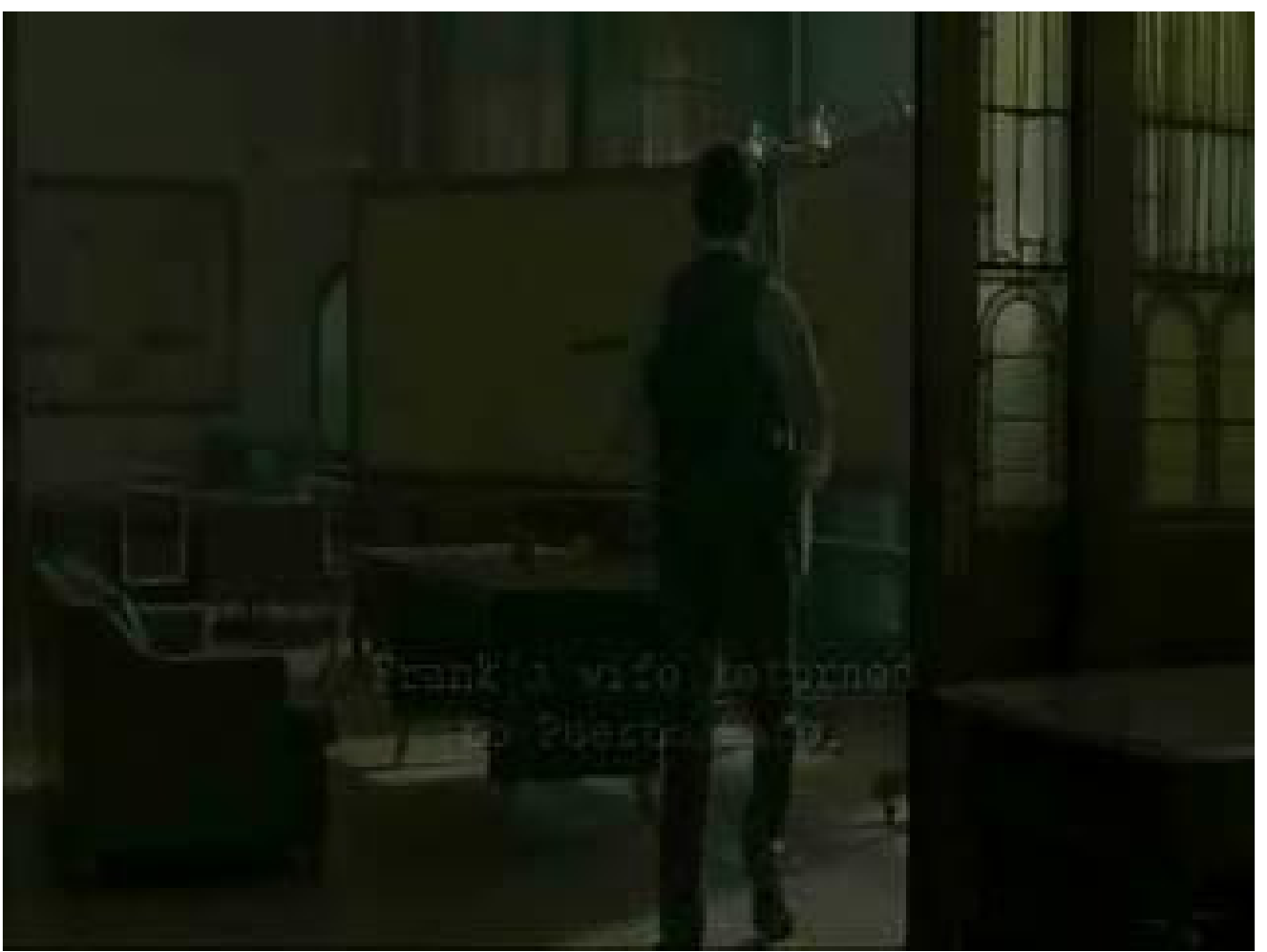}
  \hspace{-1.5mm}
  \includegraphics[width=0.121\linewidth]{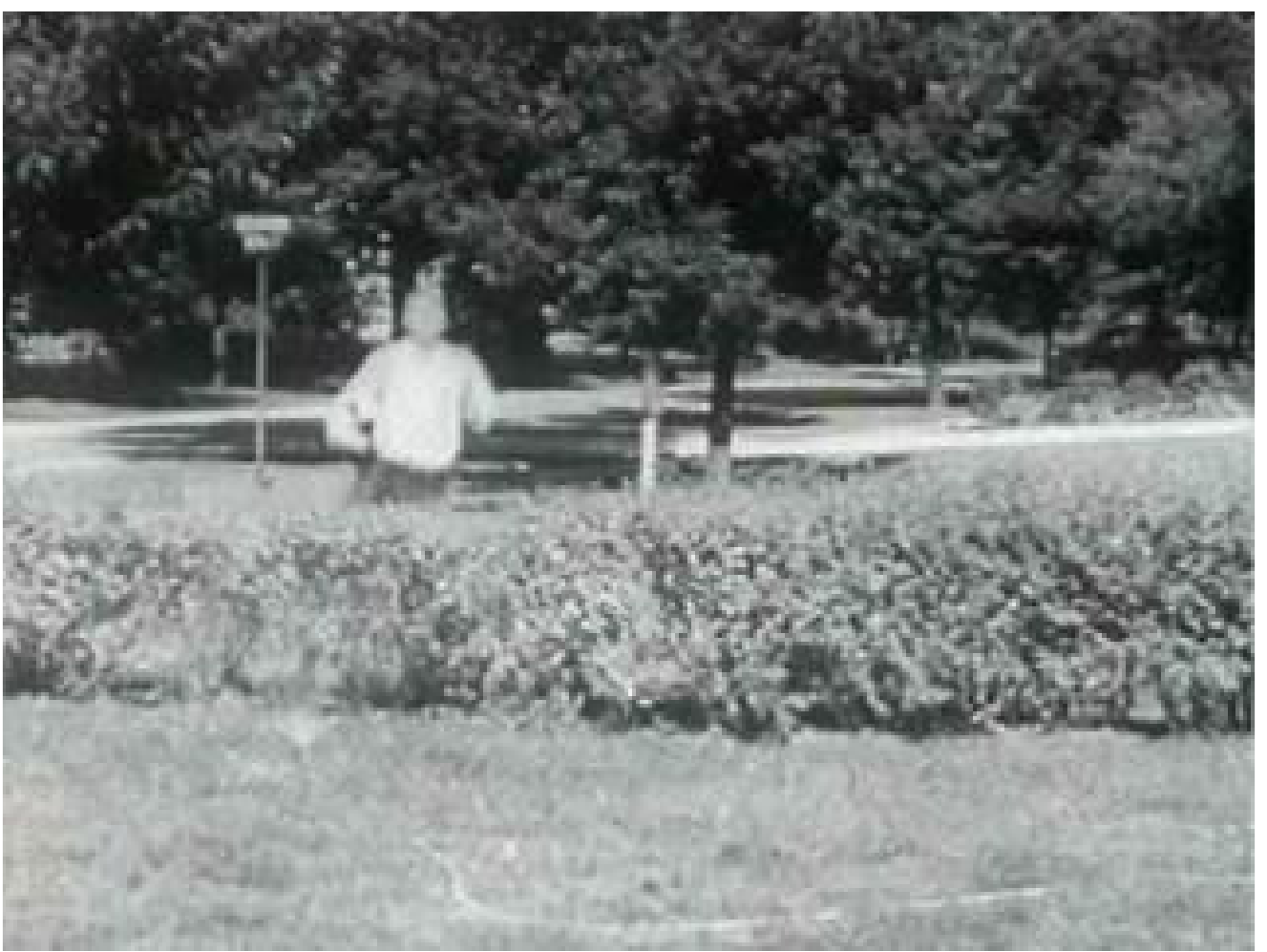}
  \includegraphics[width=0.121\linewidth]{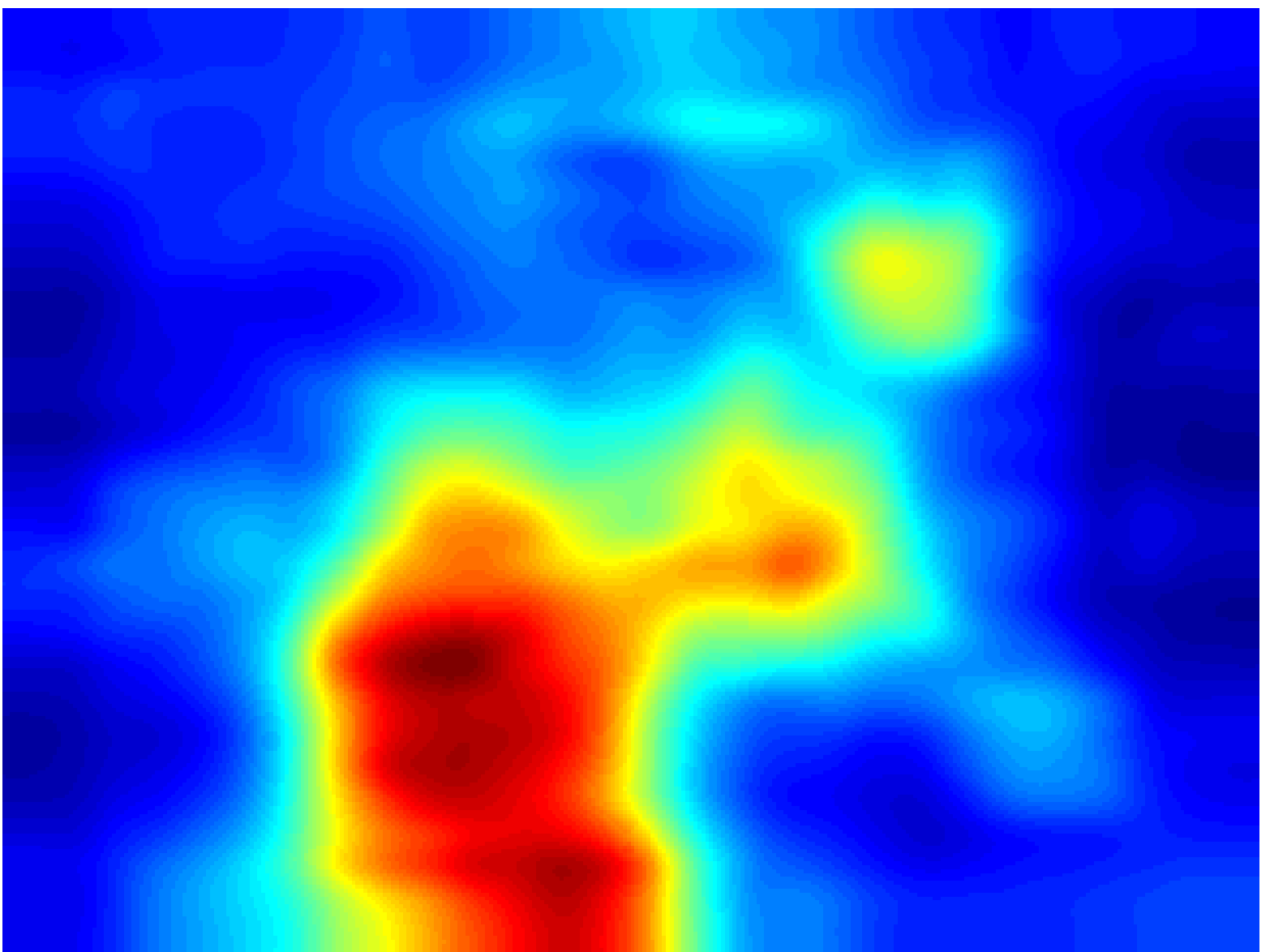}
  \hspace{-1.5mm}
  \includegraphics[width=0.121\linewidth]{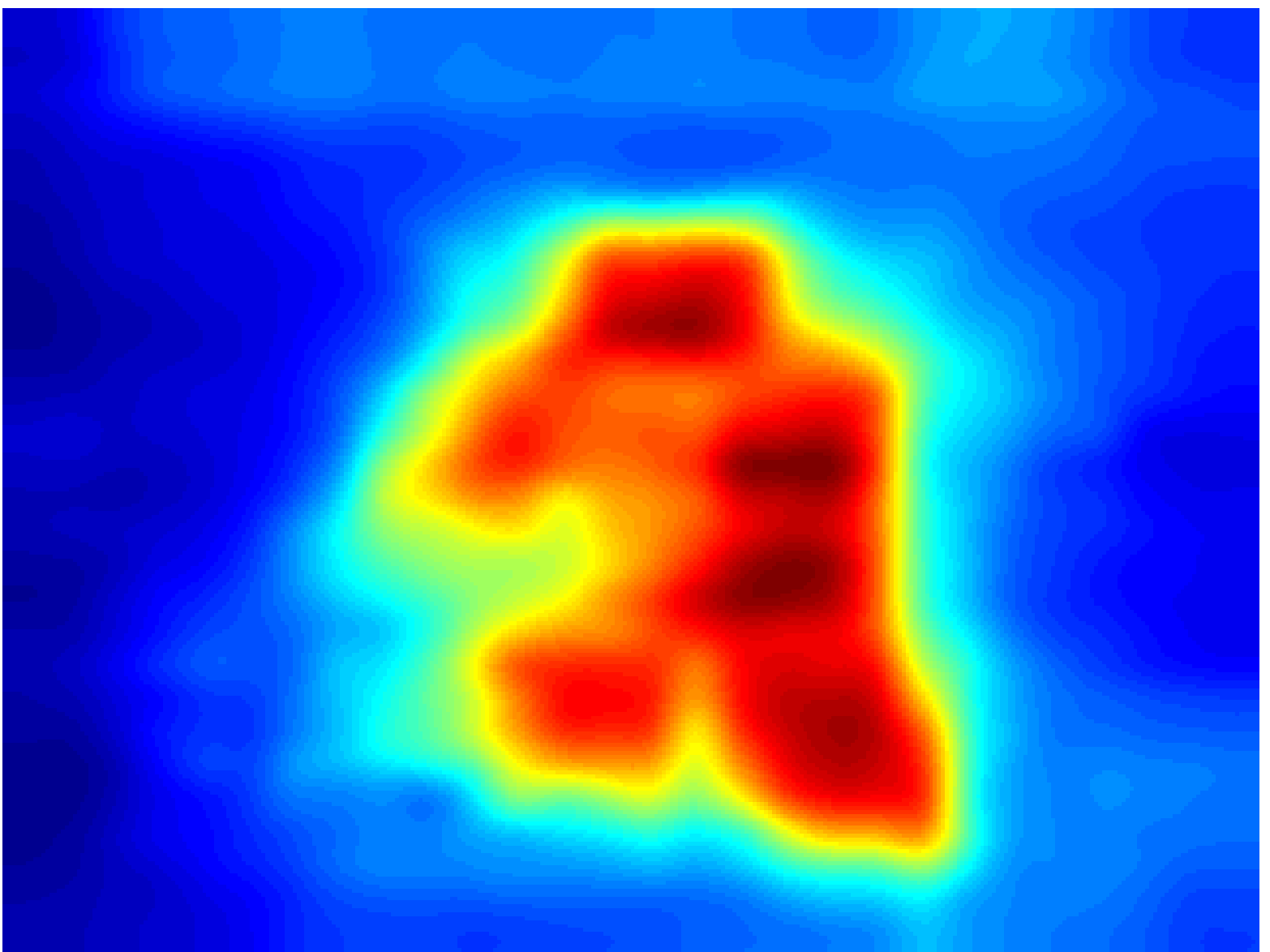}
  \hspace{-1.5mm}
  \includegraphics[width=0.121\linewidth]{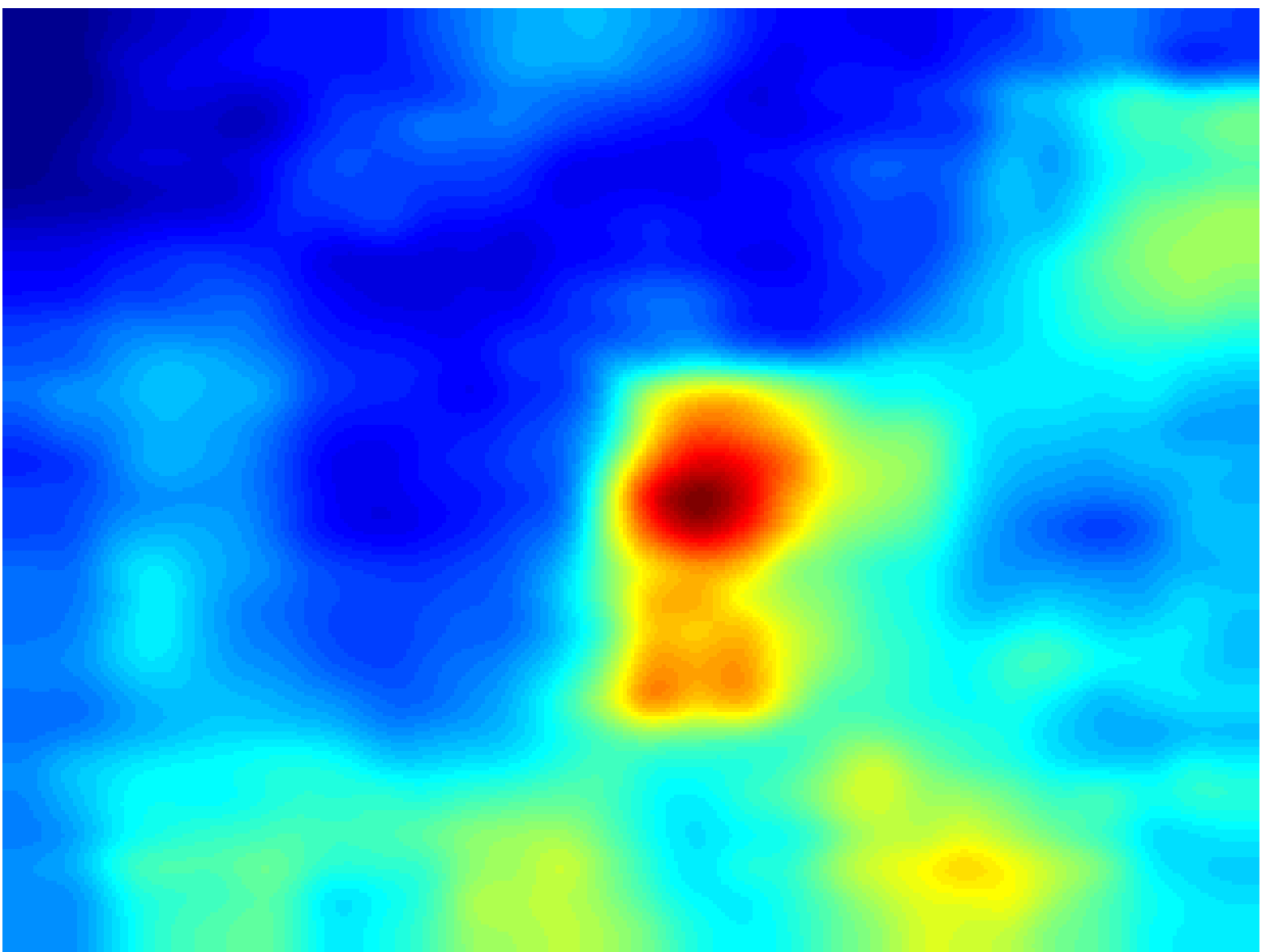}
  \hspace{-1.5mm}
  \includegraphics[width=0.121\linewidth]{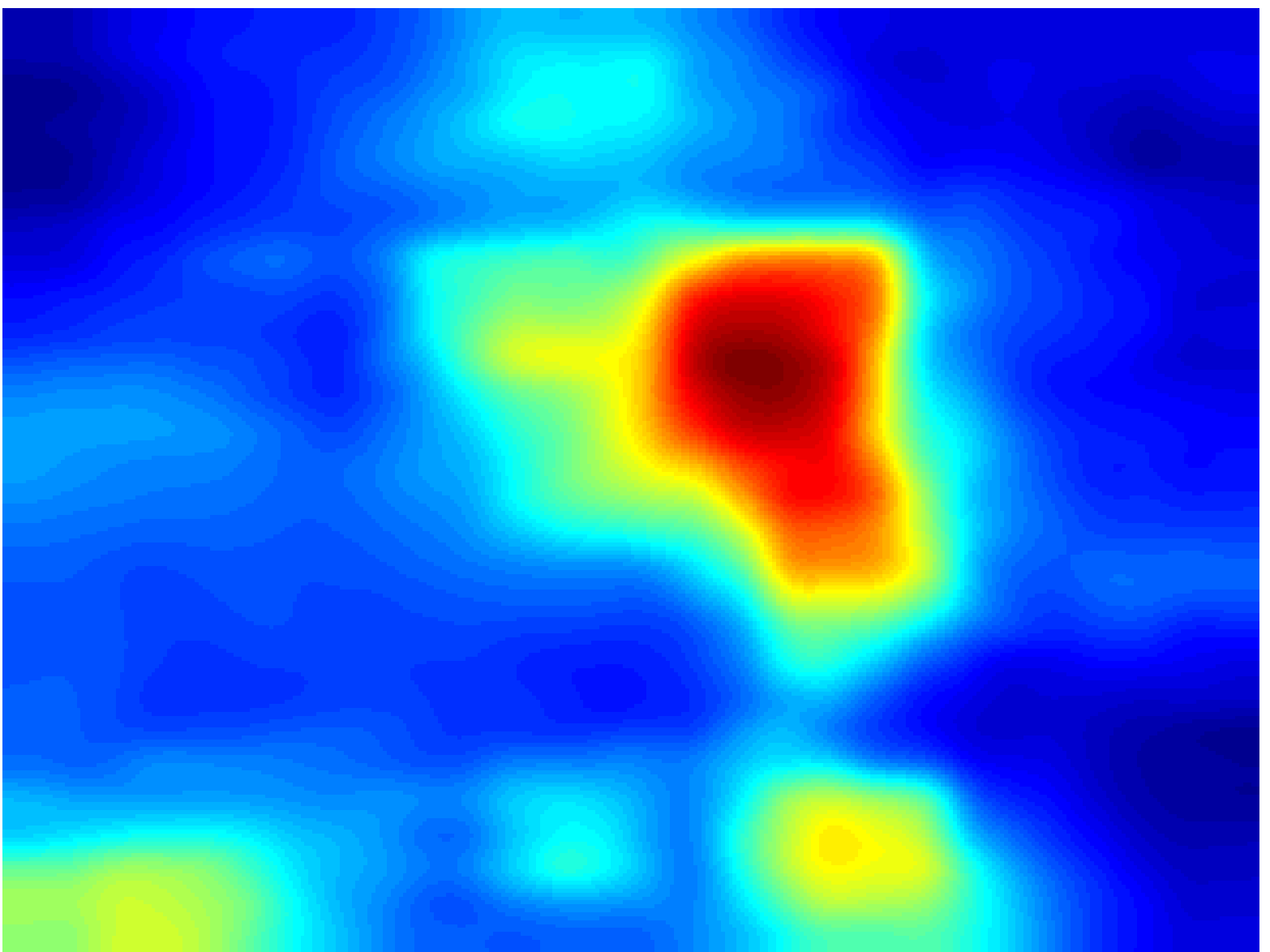}
  \hspace{-1.5mm}
  \includegraphics[width=0.121\linewidth]{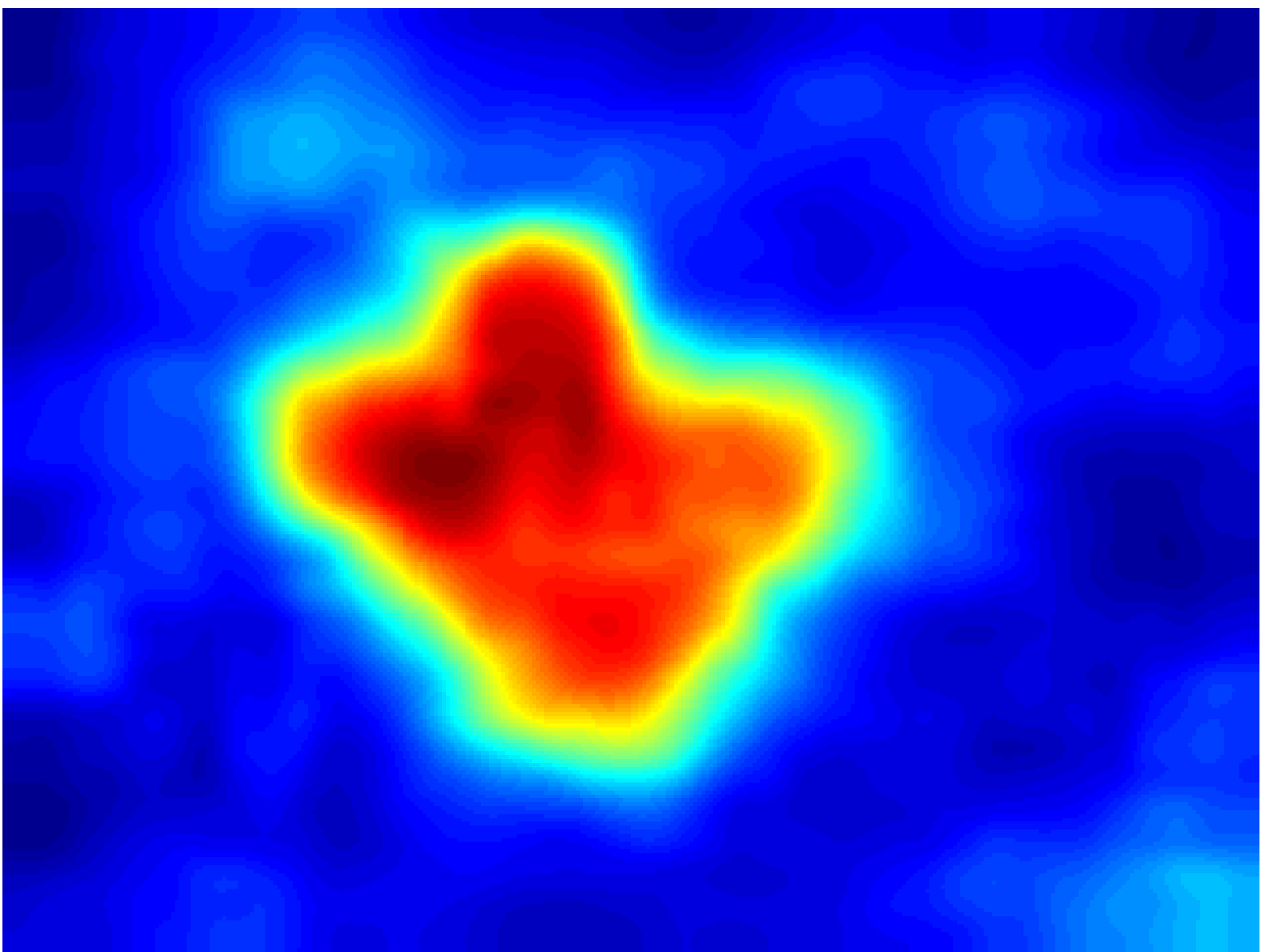}
  \hspace{-1.5mm}
  \includegraphics[width=0.121\linewidth]{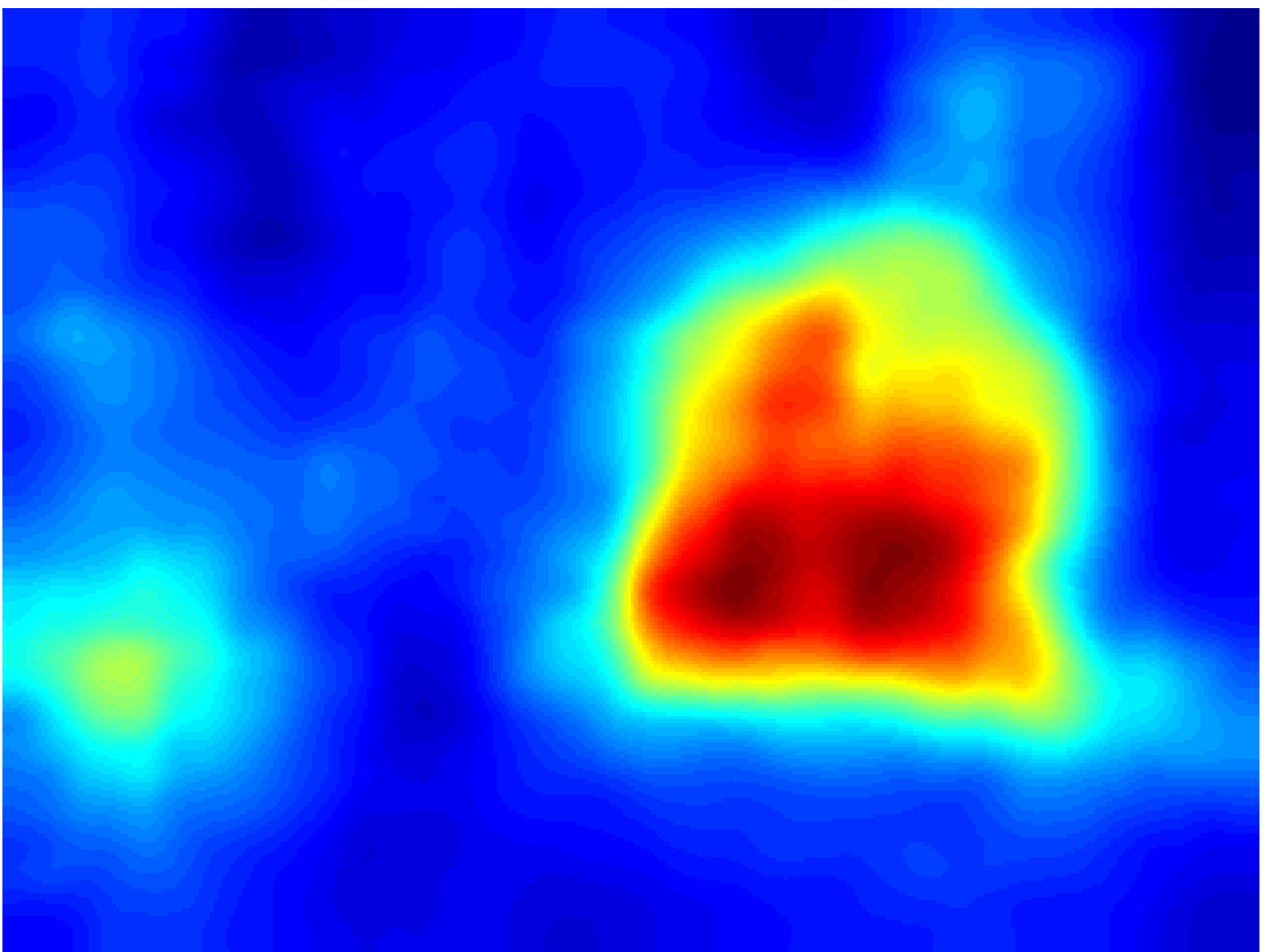}
  \hspace{-1.5mm}
  \includegraphics[width=0.121\linewidth]{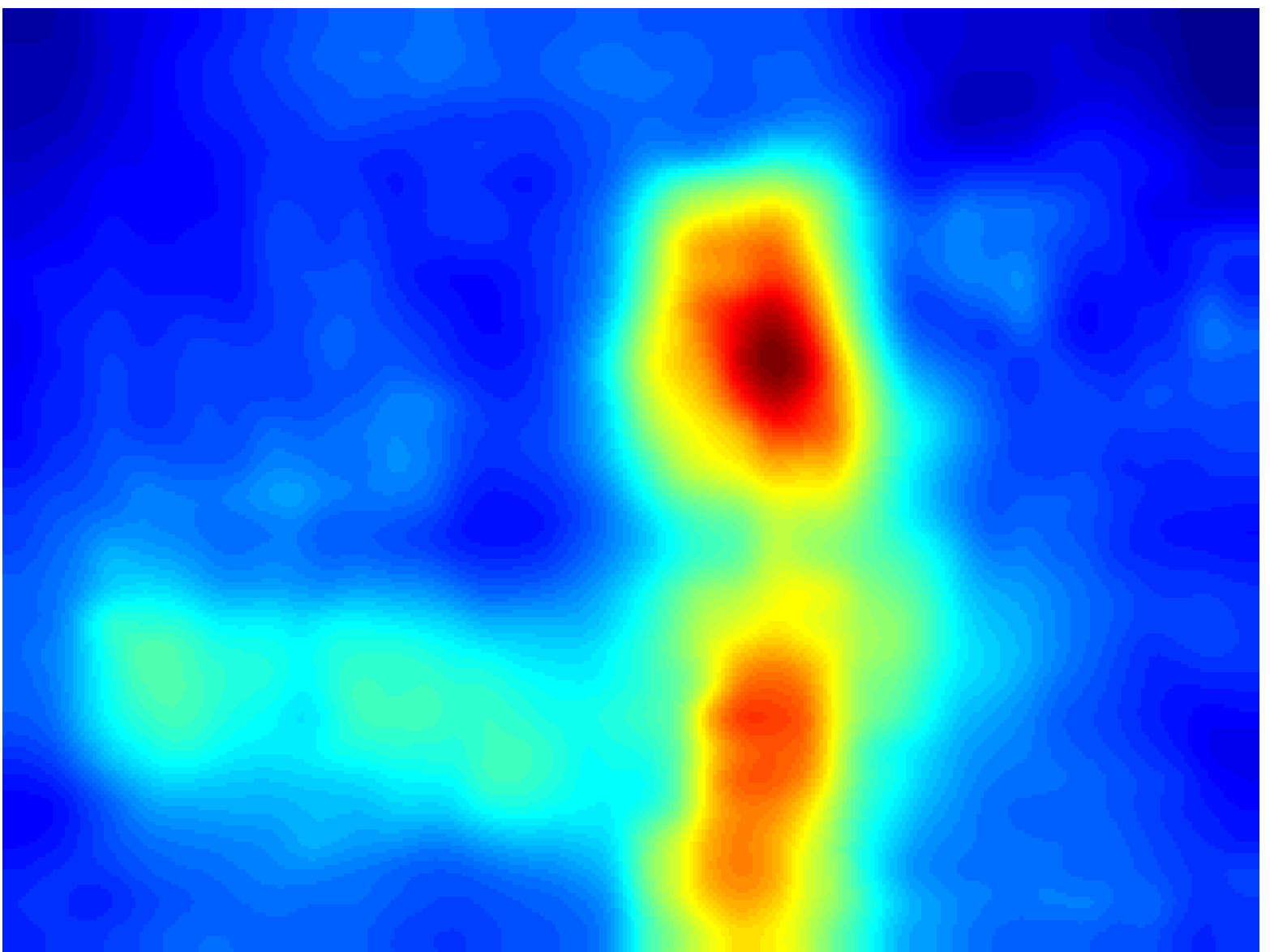}
  \hspace{-1.5mm}
  \includegraphics[width=0.121\linewidth]{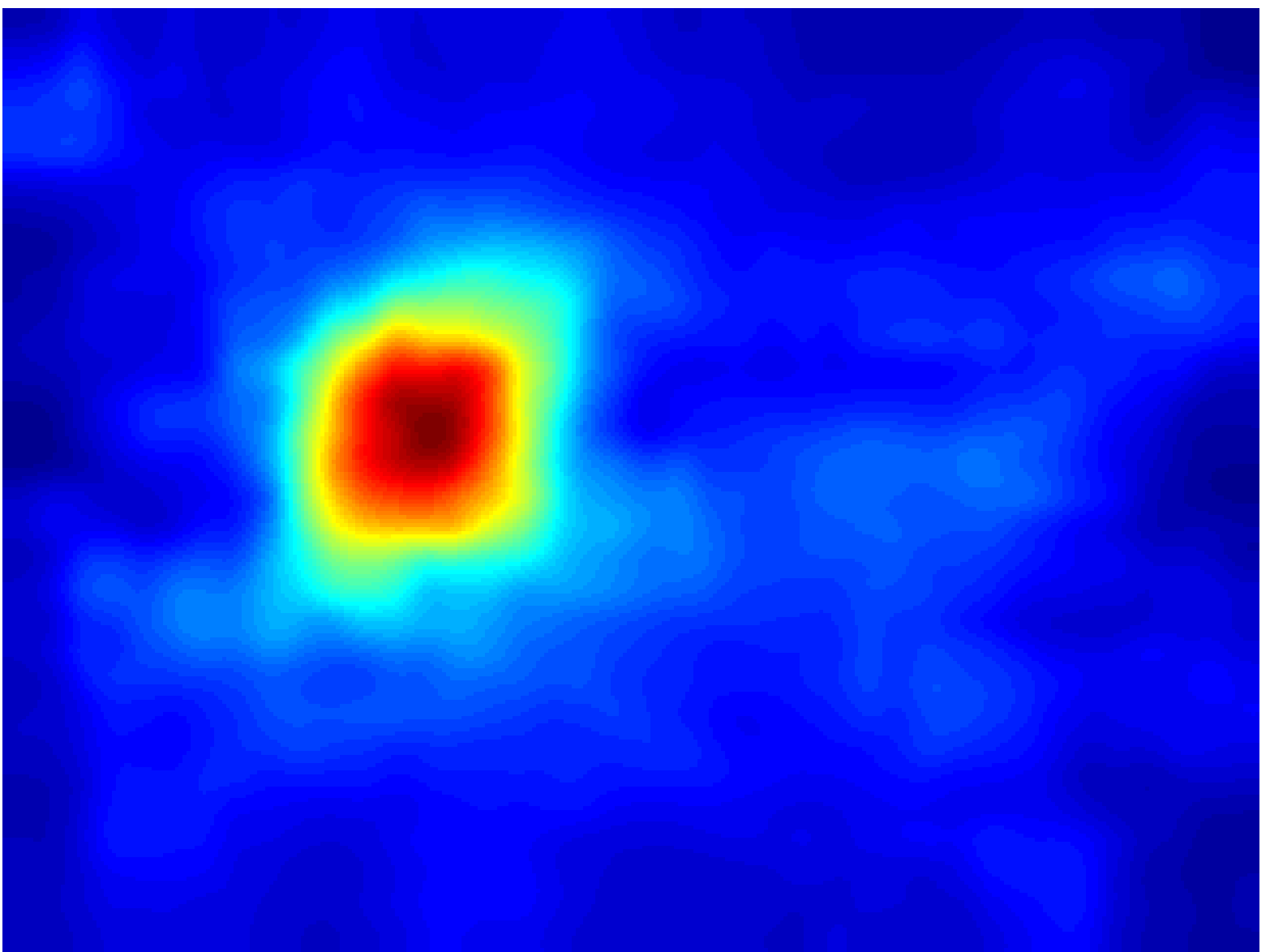}
  \includegraphics[width=0.121\linewidth]{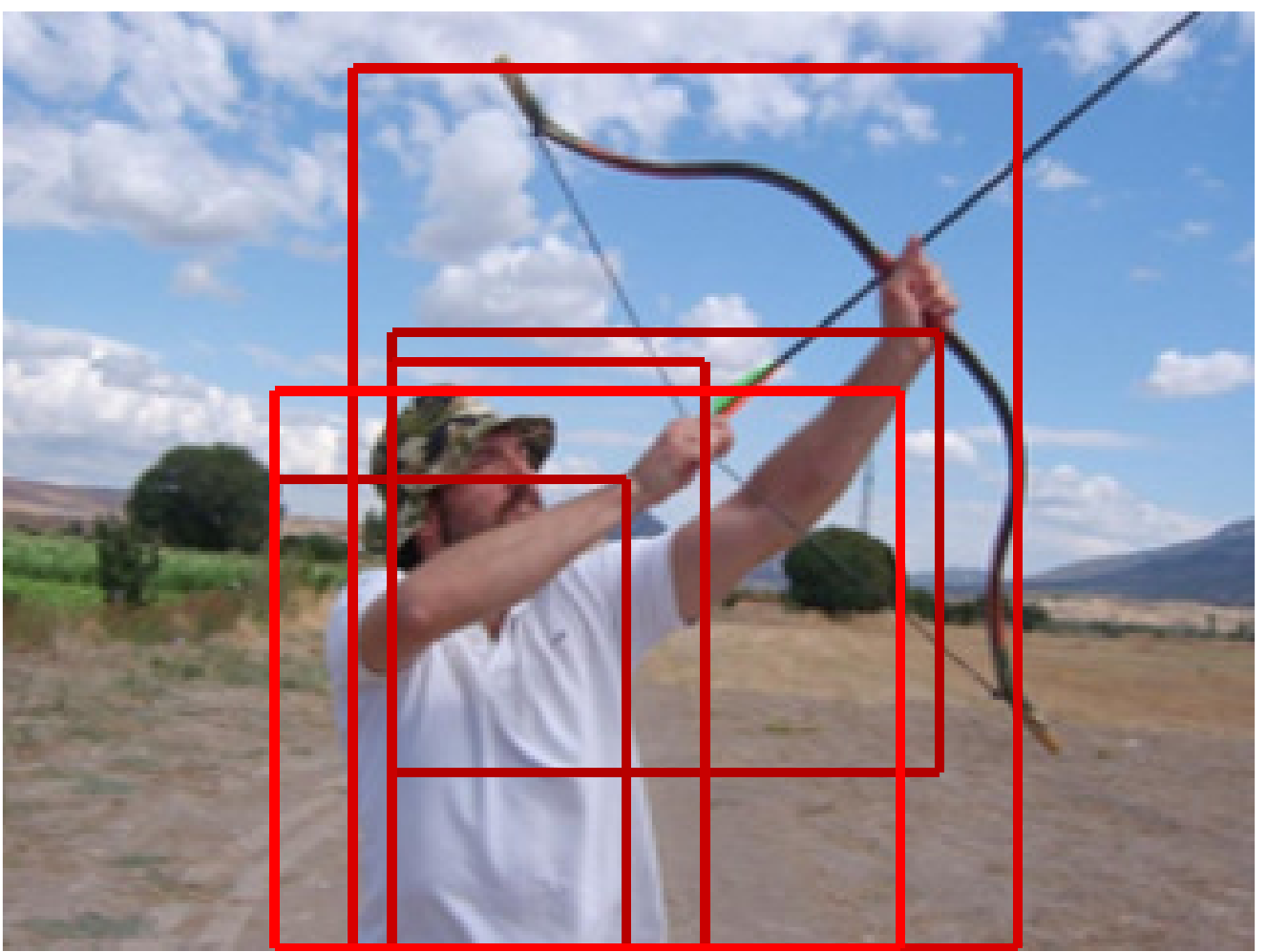}
  \hspace{-1.5mm}
  \includegraphics[width=0.121\linewidth]{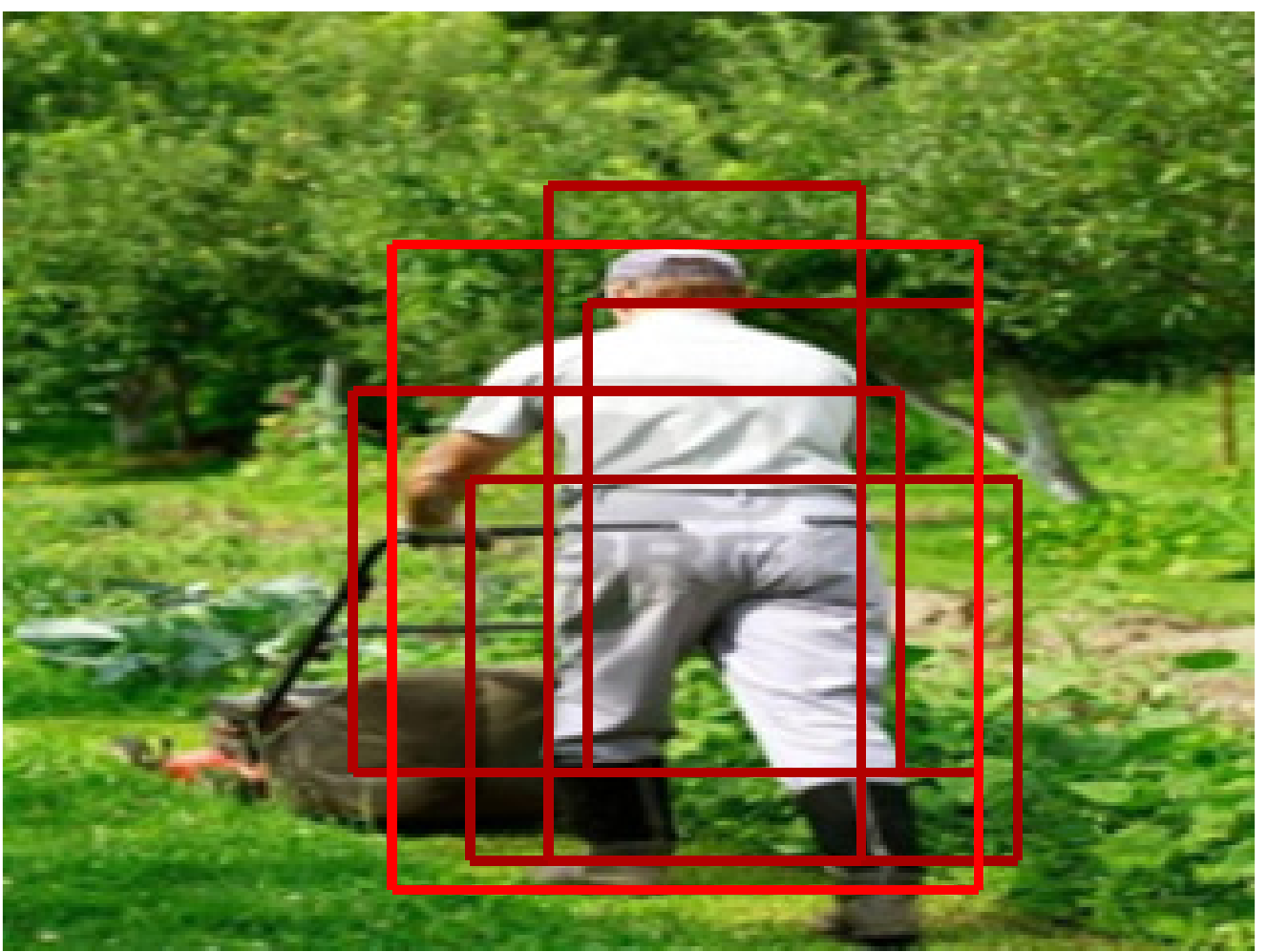}
  \hspace{-1.5mm}
  \includegraphics[width=0.121\linewidth]{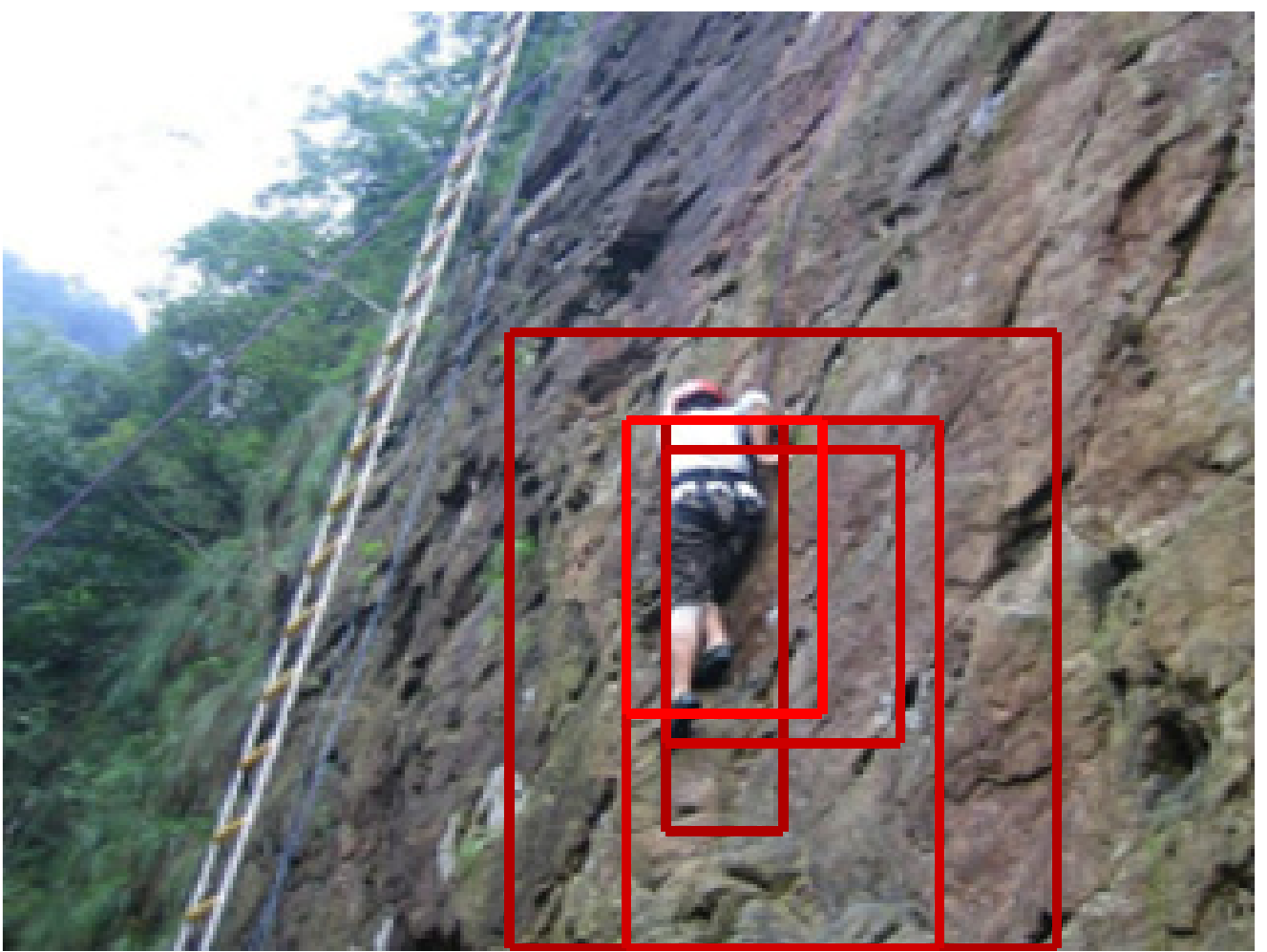}
  \hspace{-1.5mm}
  \includegraphics[width=0.121\linewidth]{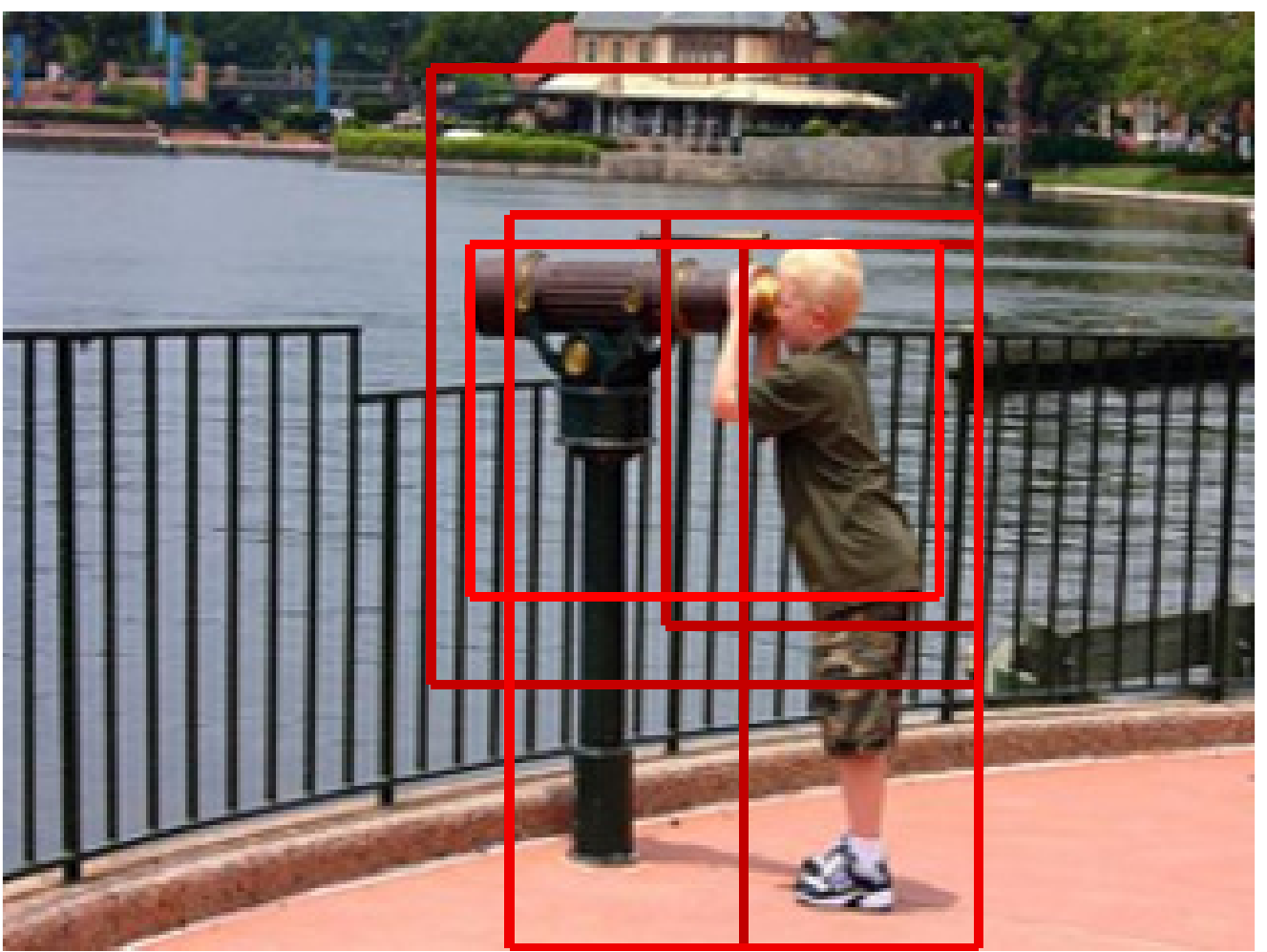}
  \hspace{-1.5mm}
  \includegraphics[width=0.121\linewidth]{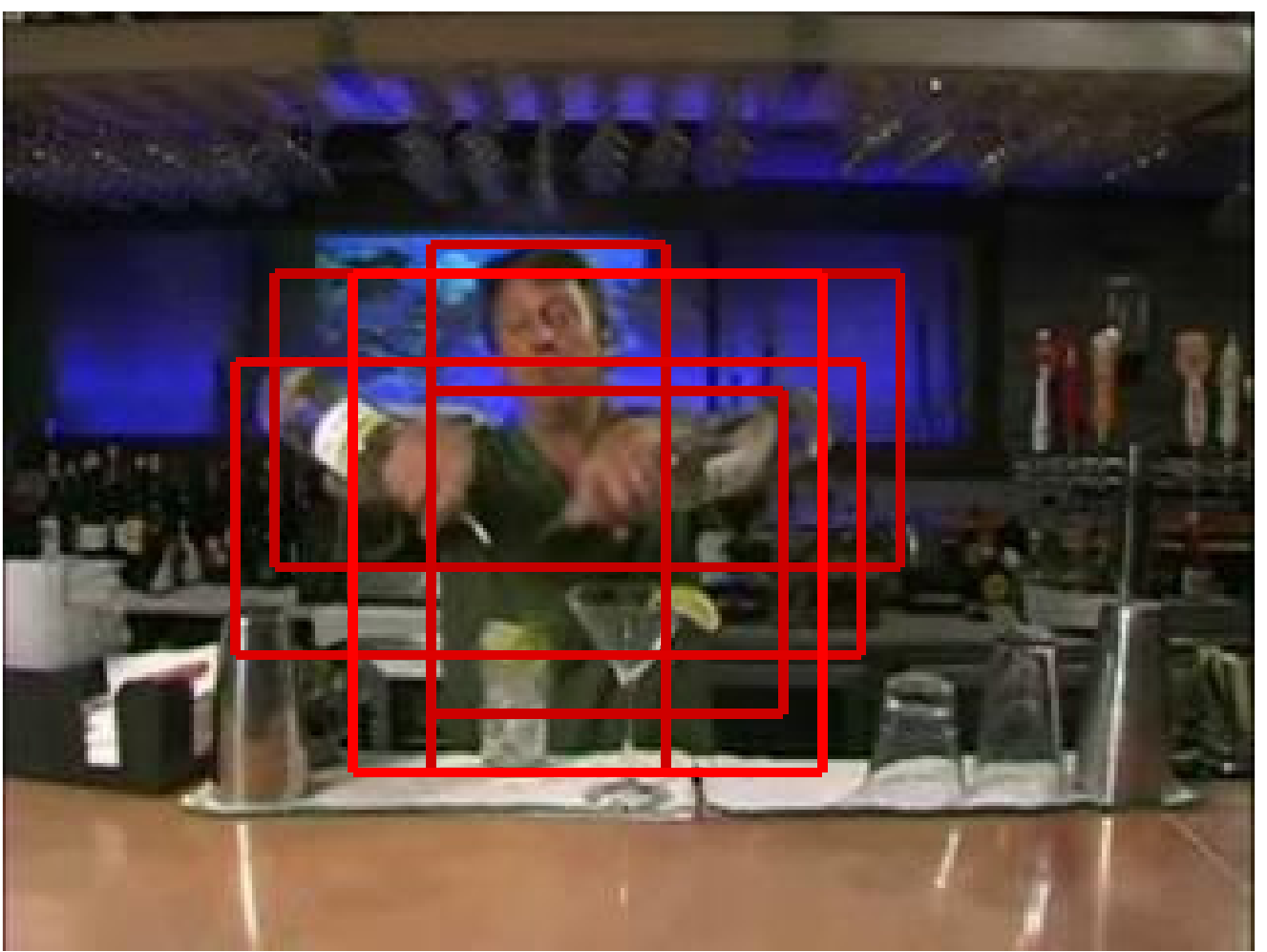}
  \hspace{-1.5mm}
  \includegraphics[width=0.121\linewidth]{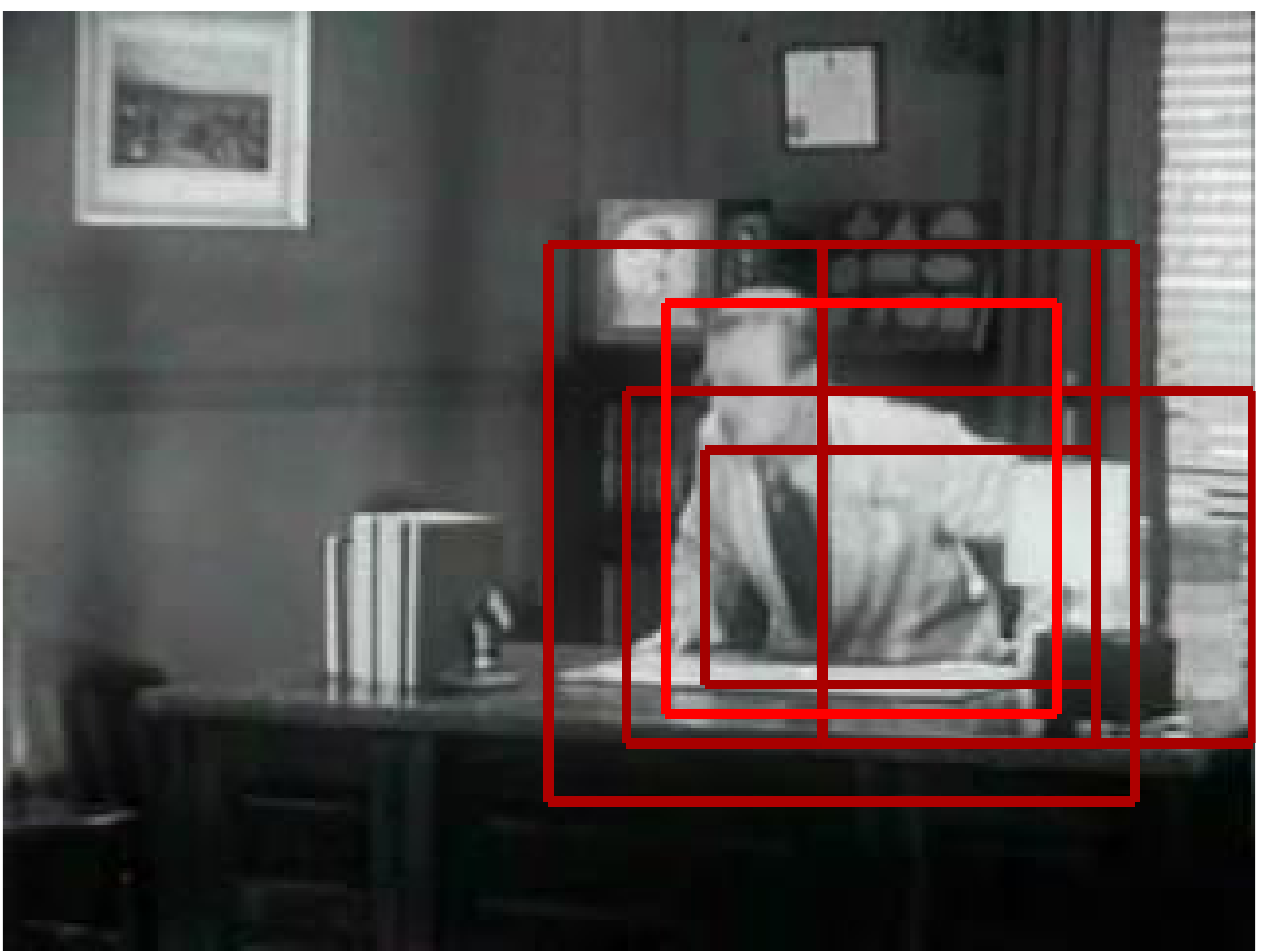}
  \hspace{-1.5mm}
  \includegraphics[width=0.121\linewidth]{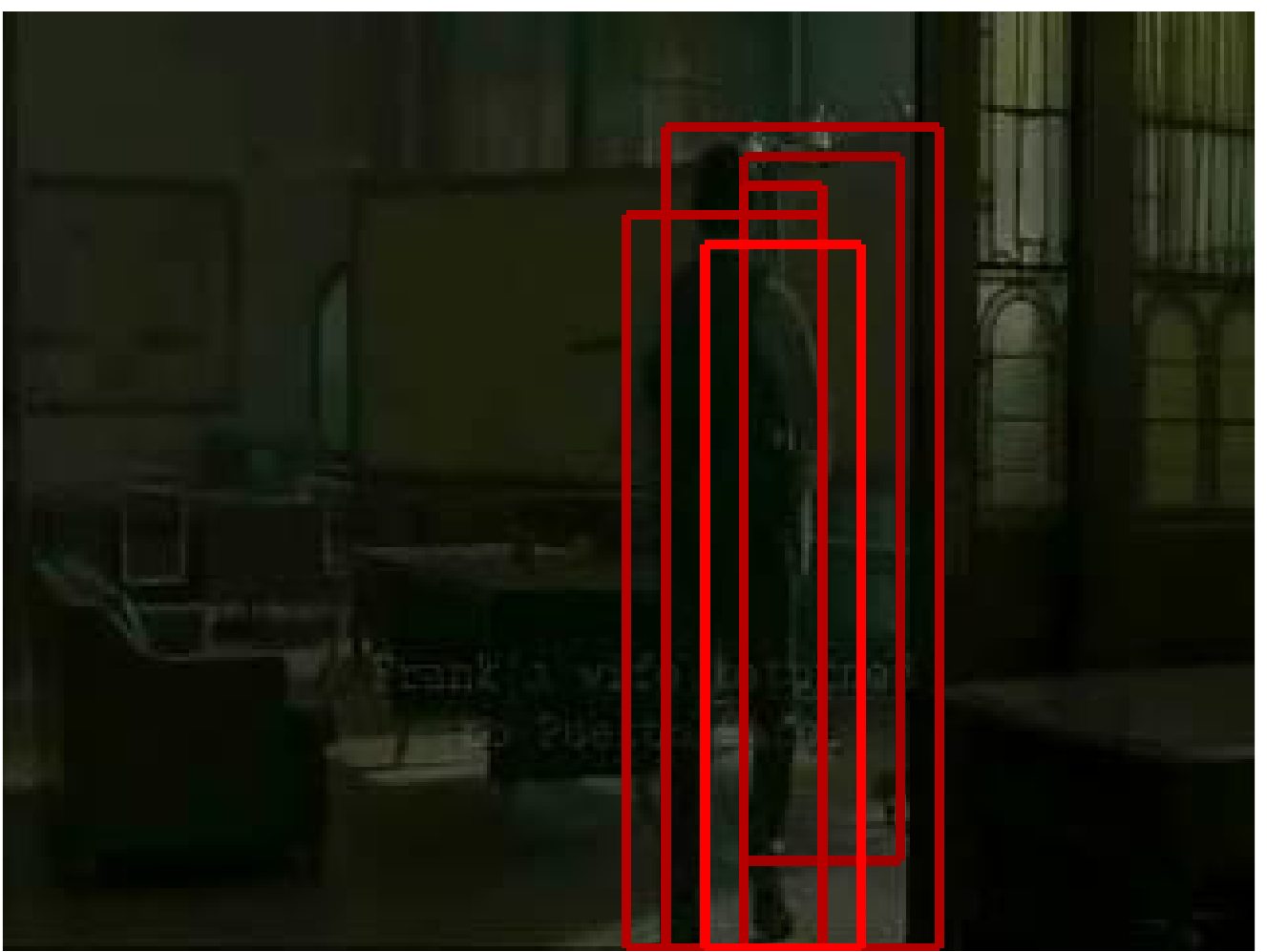}
  \hspace{-1.5mm}
  \includegraphics[width=0.121\linewidth]{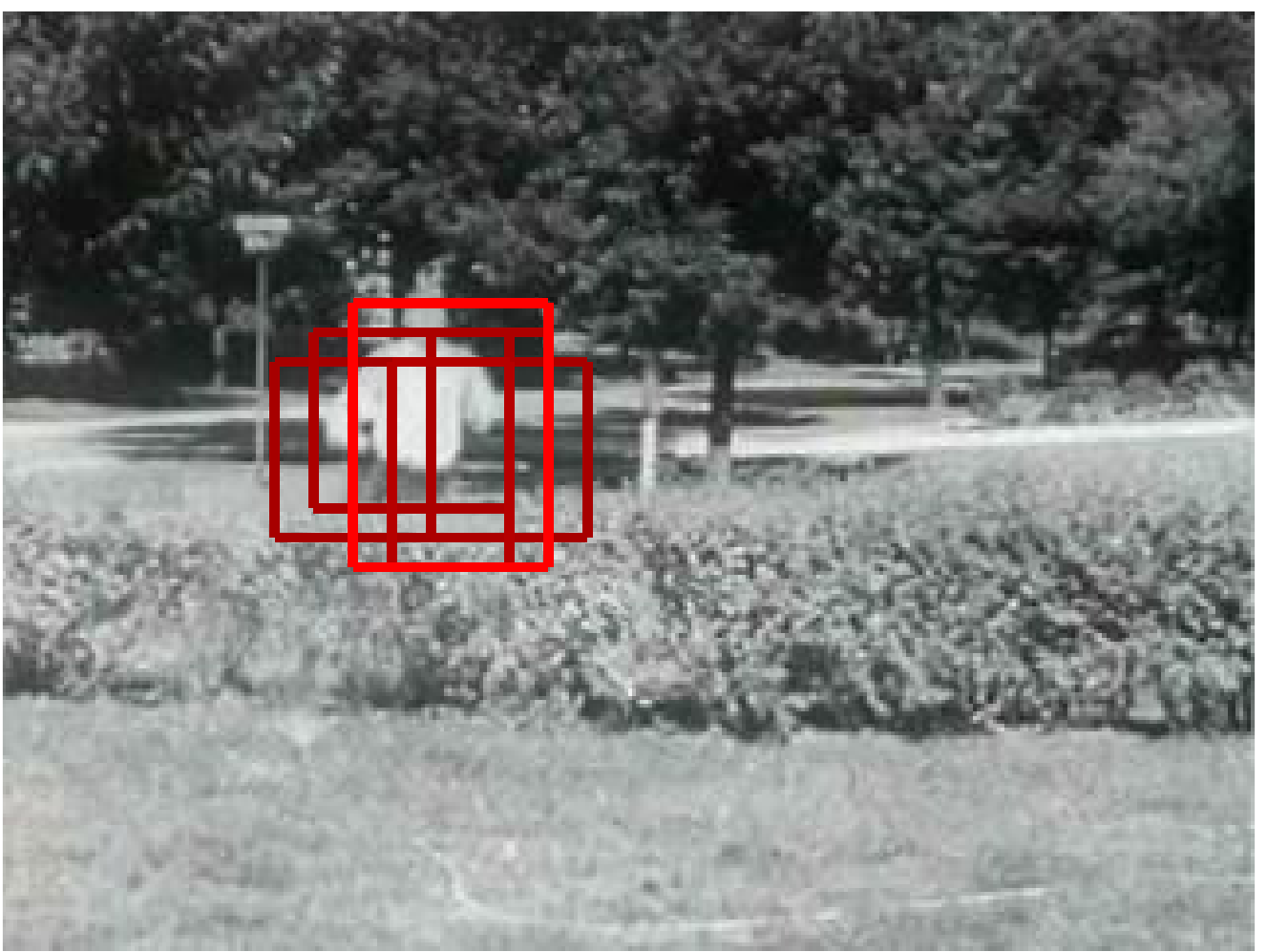}
  \caption{Examples of actionness maps and action proposals. We generate 5 action proposals for each image in this illustration. The first 4 images are from the dataset of Stanford 40 and the last 4 images are from the dataset of JHMDB. Best viewed in color.}
  \vspace{-4mm}
  \label{fig:example-actionness}
\end{figure*}

\begin{figure*}
\centering
  \subfigure[Recall at IoU above 0.5]{
  \begin{minipage}[b]{0.24\linewidth}
    \includegraphics[width=\linewidth]{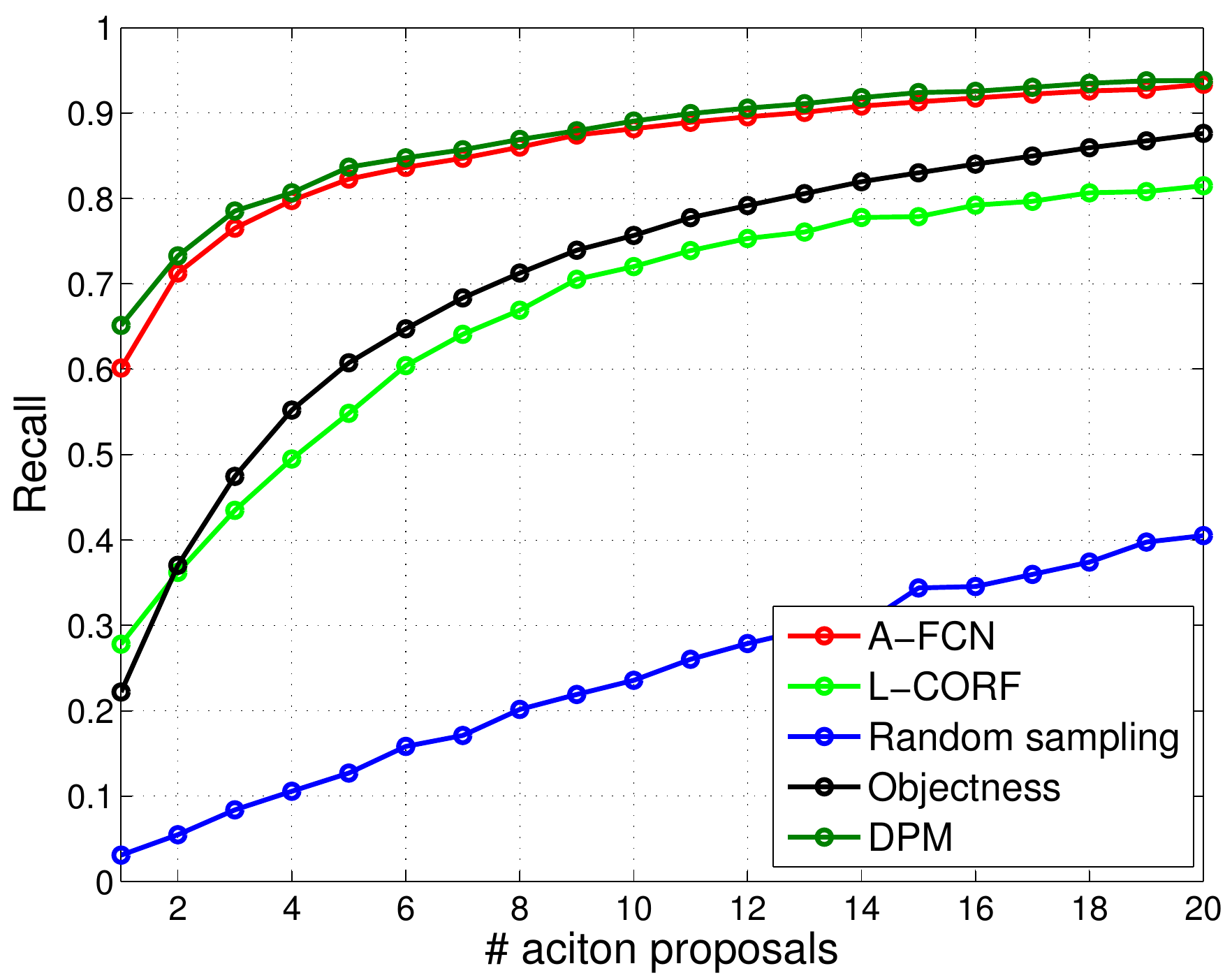}
  \end{minipage}
  }
  \hspace{-3mm}
  \subfigure[Recall at IoU above 0.7]{
  \begin{minipage}[b]{0.24\linewidth}
    \includegraphics[width=\linewidth]{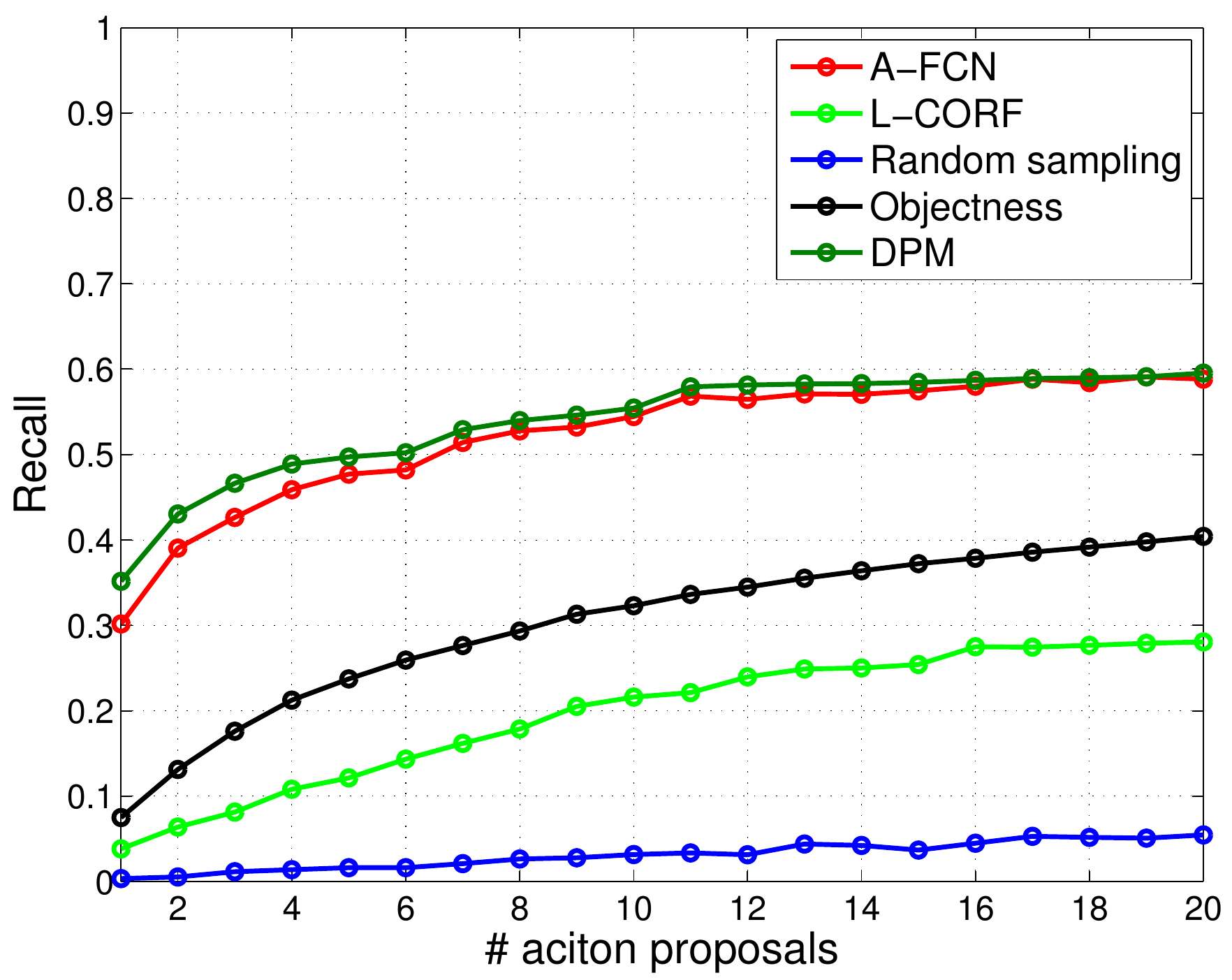}
  \end{minipage}
  }
  \hspace{-3mm}
   \subfigure[10 action proposals per image]{
  \begin{minipage}[b]{0.24\linewidth}
    \includegraphics[width=\linewidth]{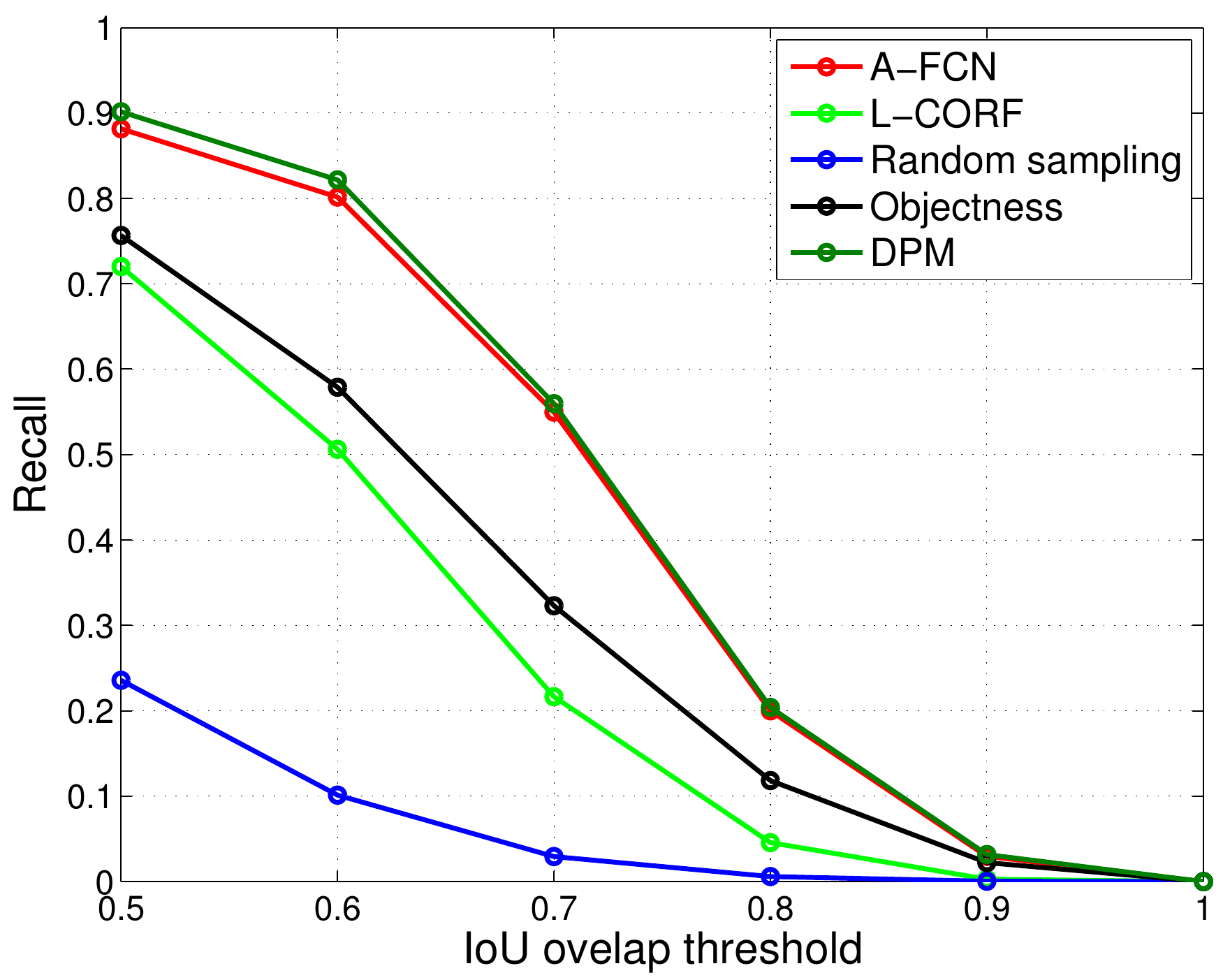}
  \end{minipage}
  }
  \hspace{-3mm}
   \subfigure[5 action proposals per image]{
  \begin{minipage}[b]{0.24\linewidth}
    \includegraphics[width=\linewidth]{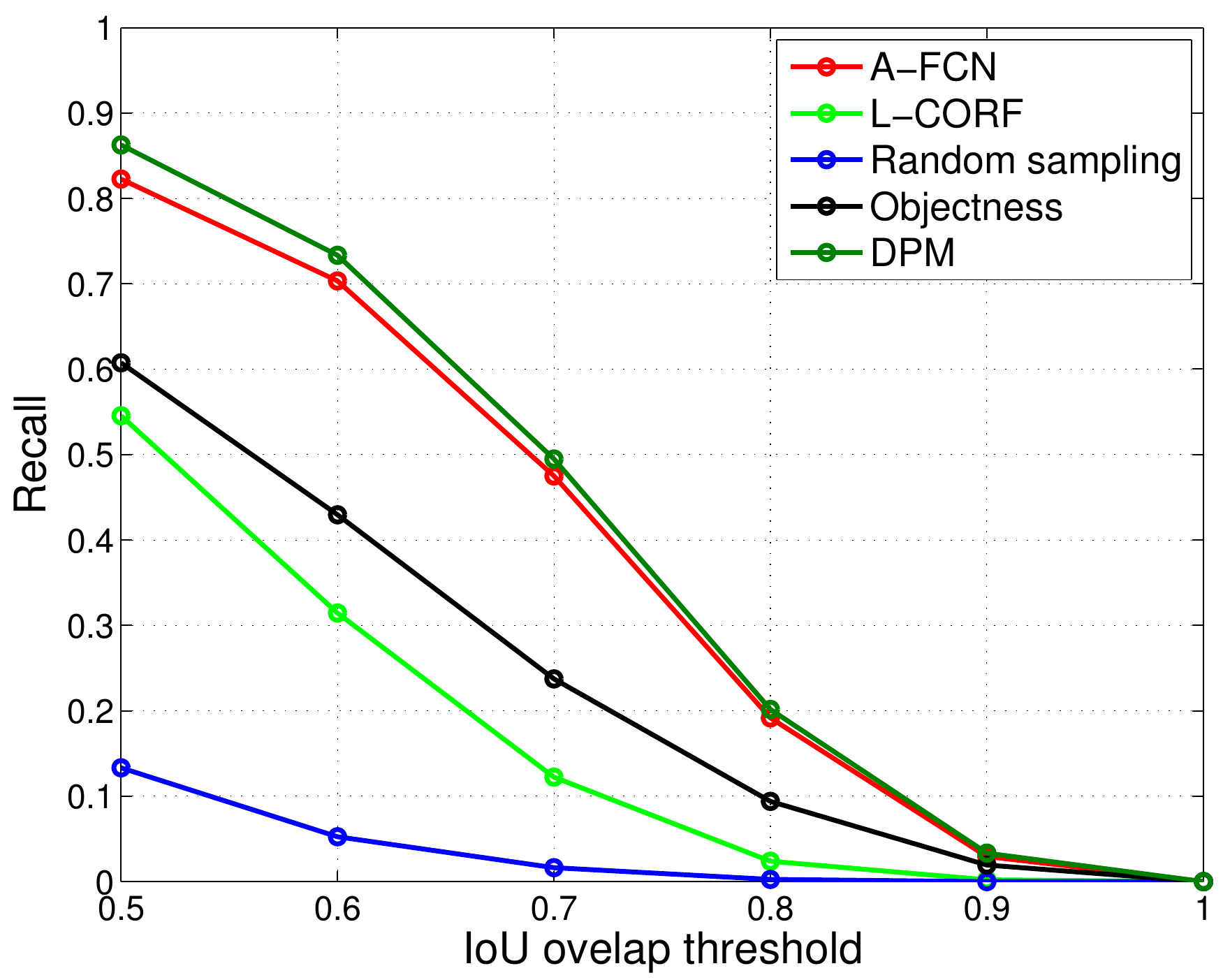}
  \end{minipage}
  } \\ \vspace{-2mm}
  \subfigure[Recall at IoU above 0.5]{
  \begin{minipage}[b]{0.24\linewidth}
    \includegraphics[width=\linewidth]{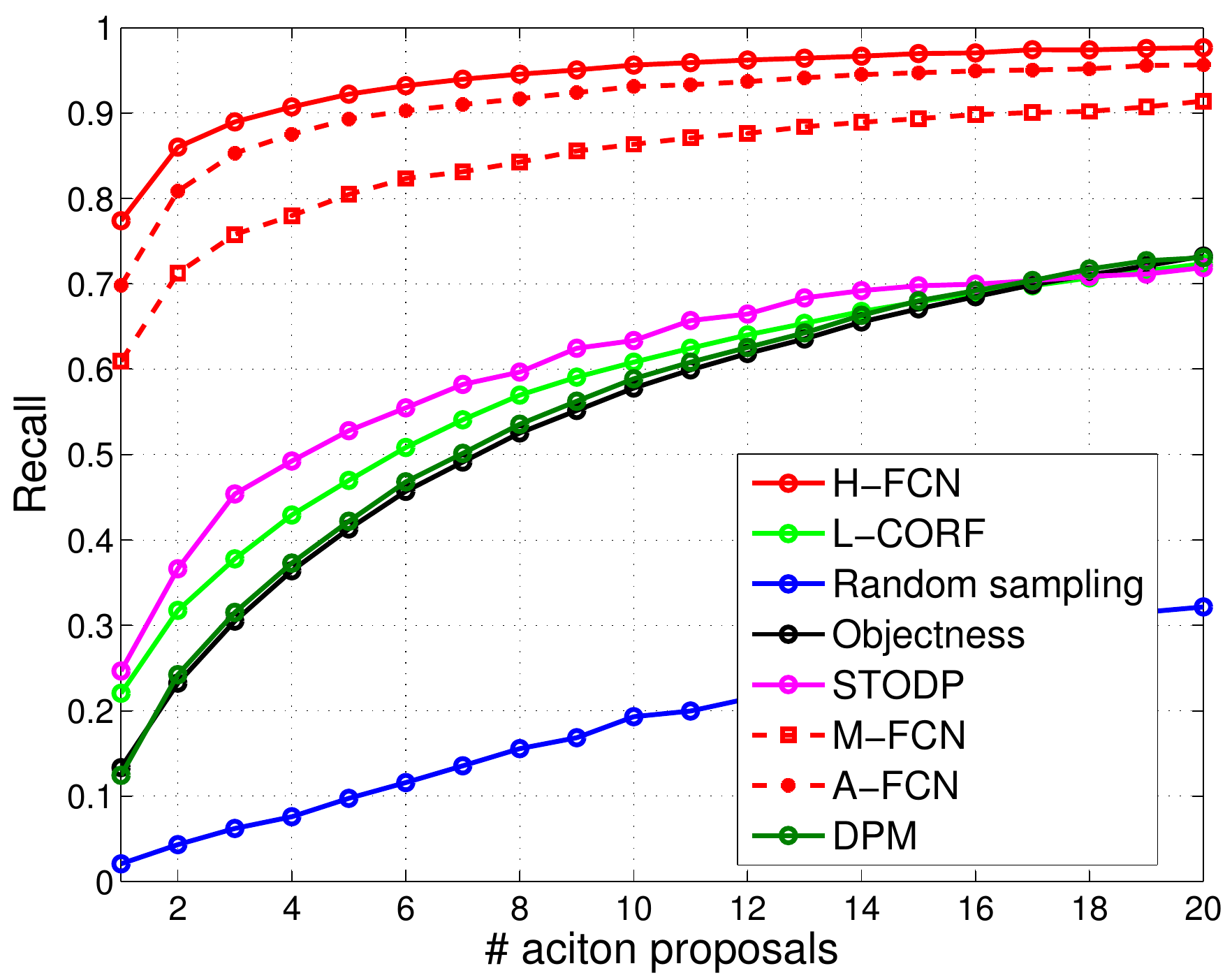}
  \end{minipage}
  }
  \hspace{-3mm}
  \subfigure[Recall at IoU above 0.7]{
  \begin{minipage}[b]{0.24\linewidth}
    \includegraphics[width=\linewidth]{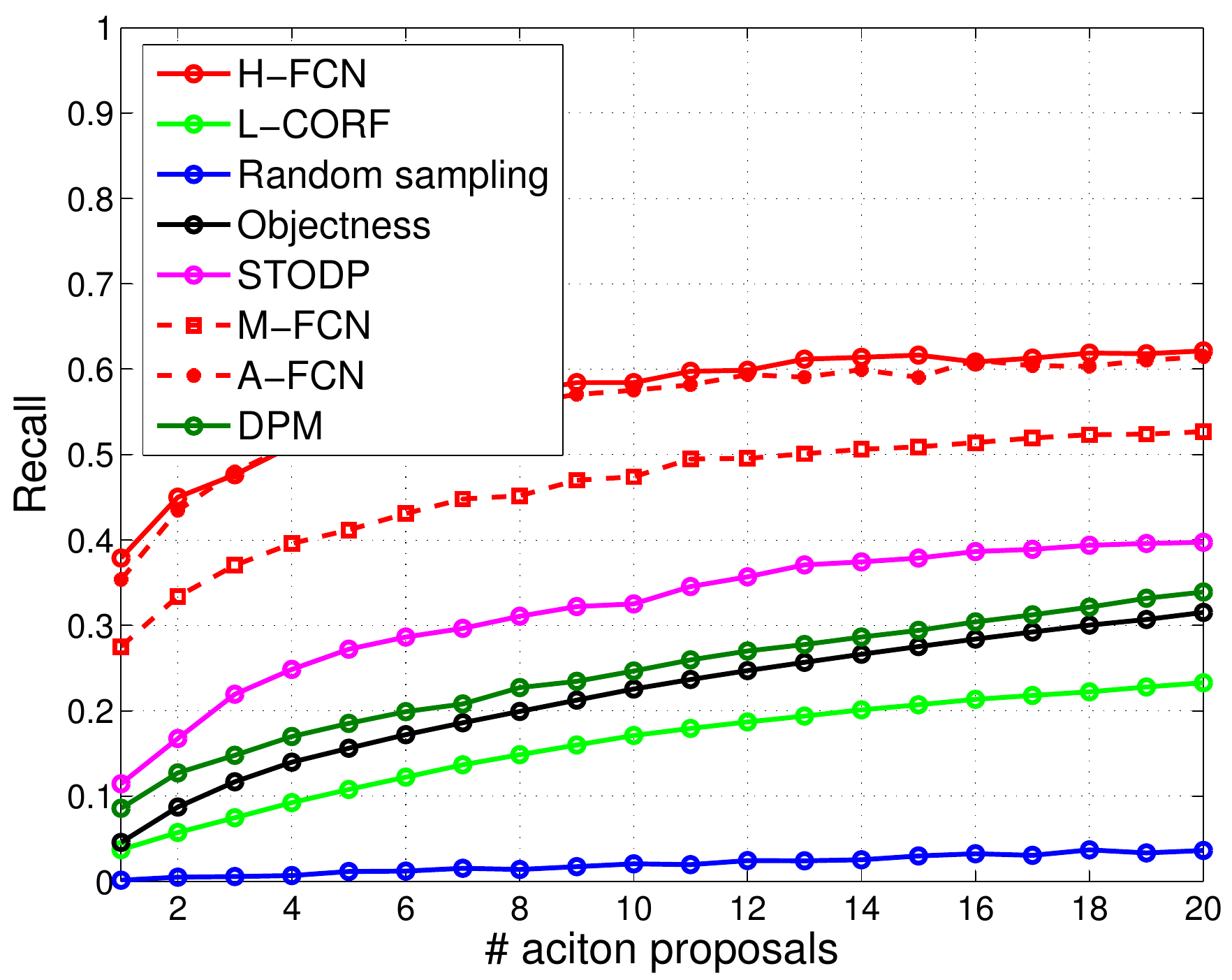}
  \end{minipage}
  }
  \hspace{-3mm}
  \subfigure[10 action proposals per image]{
  \begin{minipage}[b]{0.24\linewidth}
    \includegraphics[width=\linewidth]{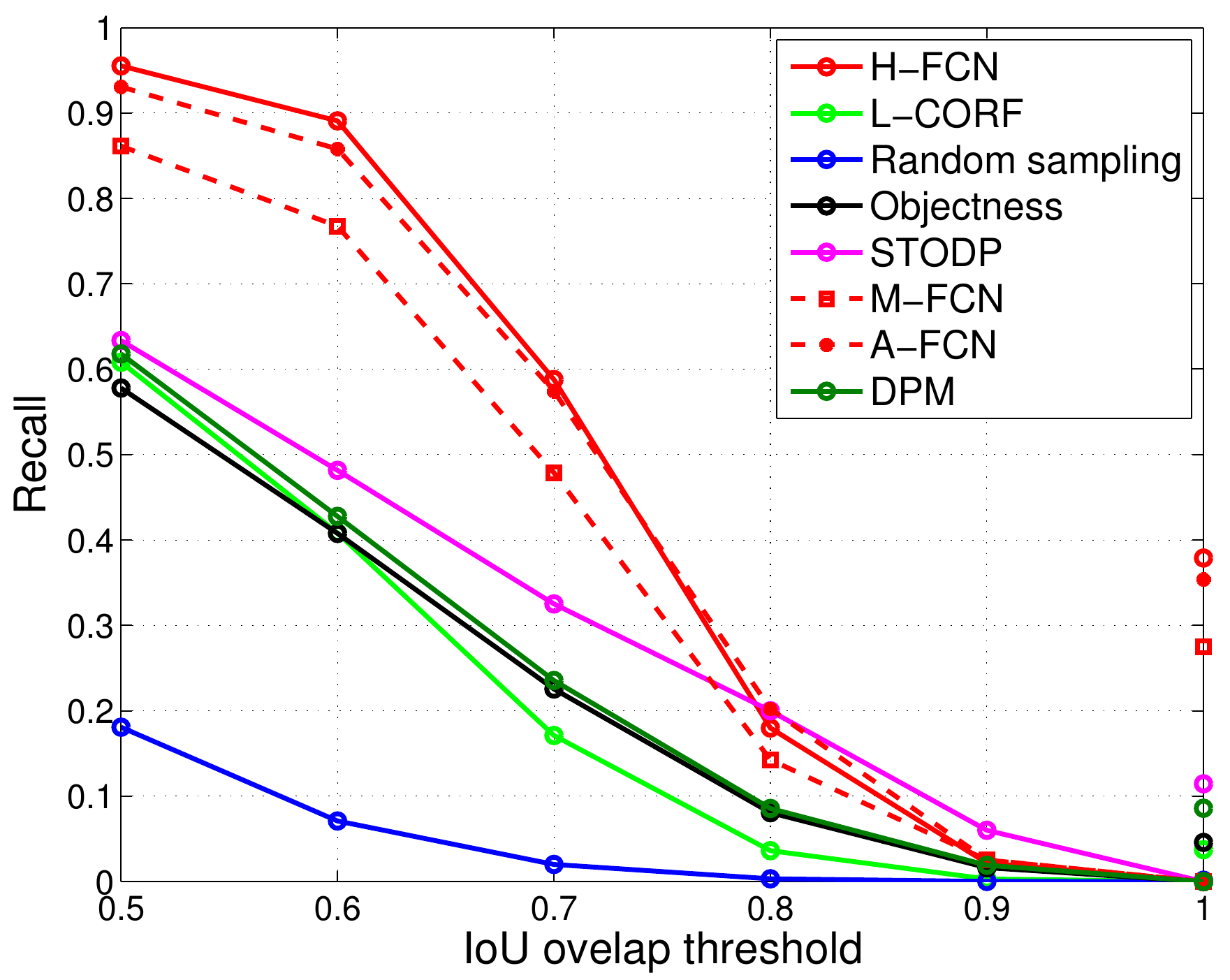}
  \end{minipage}
  }
  \hspace{-3mm}
   \subfigure[5 action proposals per image]{
  \begin{minipage}[b]{0.24\linewidth}
    \includegraphics[width=\linewidth]{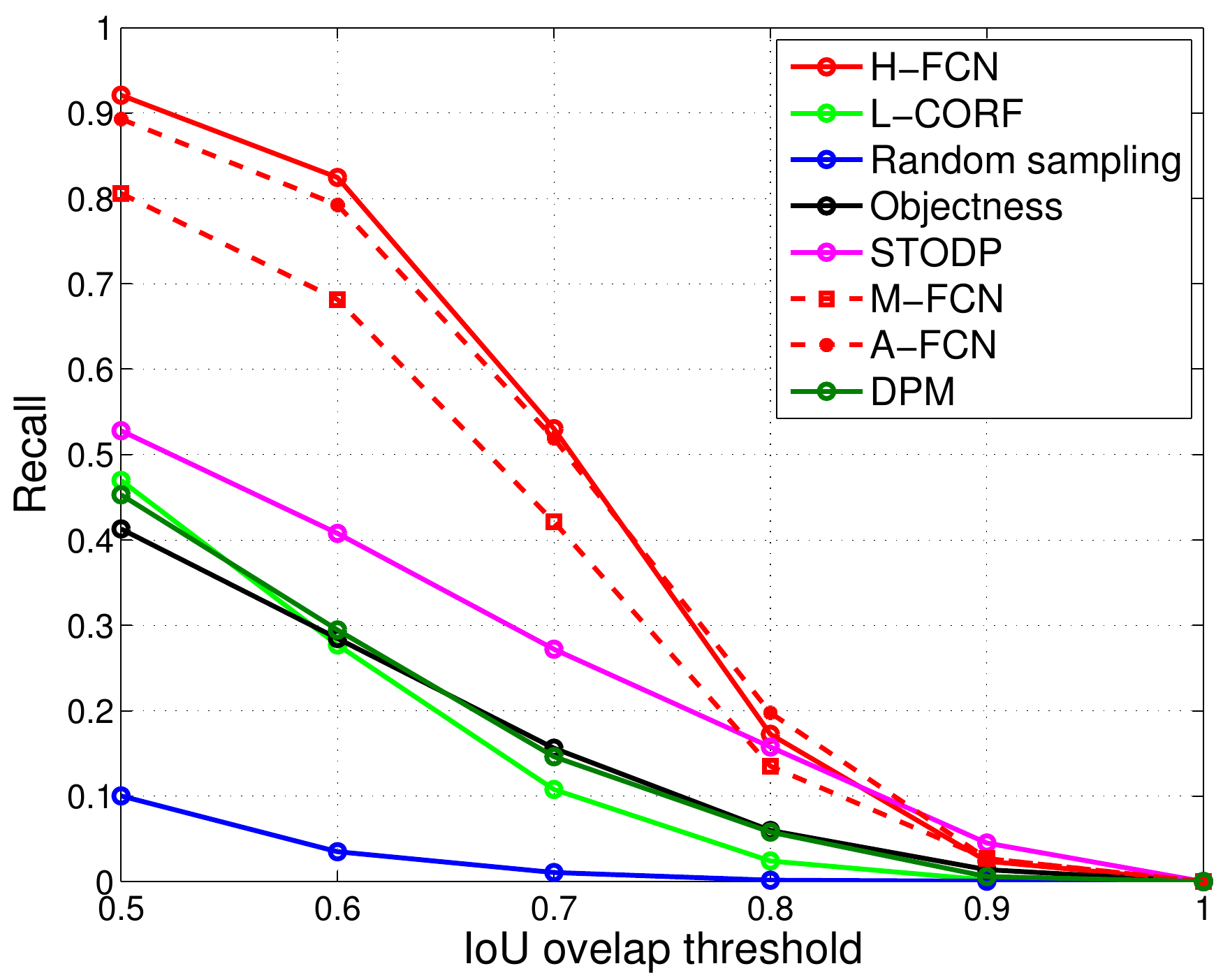}
  \end{minipage}
  }
  \caption{Evaluation of action proposals on the dataset of Stanford 40 (top row) and JHMDB (bottom row). We compare our method with previous actionness estimation approach (L-CORF) \cite{ChenXXC14}, Spatio-temporal object detection proposal (STODP) \cite{OneataRVS14}, objectness \cite{AlexeDF12},  DPM \cite{FelzenszwalbGMR10} and random sampling. Best viewed in color.}
  \label{fig:actionproposal}
  \vspace{-4mm}
\end{figure*}

\begin{table*}
  \resizebox{\textwidth}{!}{
  \begin{tabular}{l|ccccccccccccccccccccc|l}
    \textbf{frame-AP}(\%)& brush-hair & catch& clap & climb & golf & jump & kick-ball & pick & pour&pullup&push&run&shoot-ball&shoot-bow&shoot-gun&sit&stand&swing-baseball&throw&walk&wave&mAP \\
    \hline
    spatial-CNN \cite{GkioxariM14} & 55.8 & 25.5 & 25.1 & 24.0 & 77.5 & 1.9 & 5.3 & 21.4 & 68.6 & 71.0 & 15.4 & 6.3 & 4.6 & 41.1 & 28.0 & 9.4 & 8.2 & 19.9 & 17.8 & 29.2 & 11.5 & 27.0 \\
    motion-CNN \cite{GkioxariM14} & 32.3 & 5.0 & 35.6 & 30.1 & 58.0 & 7.8 & 2.6 & 16.4 & 55.0 & 72.3 & 8.5 & 6.1 & 3.9 & 47.8 & 7.3 & 24.9 & 26.3 & 36.3 & 4.5 & 22.1 & 7.6 & 24.3 \\
    full \cite{GkioxariM14}  & \textbf{65.2} & 18.3 & 38.1 & \textbf{39.0} & 79.4 & 7.3 & 9.4 & 25.2 & \textbf{80.2} & \textbf{82.8} & \textbf{33.6} & 11.6 & 5.6 & 66.8 & 27.0 & \textbf{32.1} & 34.2 & 33.6 & 15.5 & \textbf{34.0} & \textbf{21.9} & 36.2 \\
    \hline
    our s-net & 56.5 & 34.7 & 40.1 & 43.1 & 76.9 & 2.7 & 17.7 & 15.6 & 71.2 & 51.5 & 17.9 & 12.4 & 12.9 & 65.4 & 53.3 & 5.3 & 16.4 & 22.6 & 27.6 & 13.2 & 15.3 & 32.5 \\
    our t-net & 42.9 & 19.0 & 49.6 & 28.9 & 71.8 & 14.0 & 20.4 & 36.6 & 60.1 & 66.0 &18.0 & 17.3& 8.3 & 73.5 & 26.0 & 11.6 & 44.1 & 53.7 & 17.6 & 22.4 & 11.5 & 34.0 \\
    our full net & 60.1 & \textbf{34.2} & \textbf{56.4} & 38.9 & \textbf{83.1} & \textbf{10.8} & \textbf{24.5} & \textbf{38.5} & 71.5 & 67.5 & 21.3 & \textbf{19.8} & \textbf{11.6} & \textbf{78.0} & \textbf{50.6} & 10.9 & \textbf{43.0} & \textbf{48.9} & \textbf{26.5} & 25.2 & 15.8 & \textbf{39.9} \\
    \hline

    \textbf{video-AP}(\%) \\
    \hline
    spatial-CNN \cite{GkioxariM14} & 67.1 & 34.4 & 37.2 & 36.3 & 93.8 & 7.3 & 14.4 & 29.6 & 80.2 & 93.9 & 17.4 & 10.0 & 8.8 & 71.2 & 45.8 & 17.7 & 11.6 & 38.5 & 20.4 & 40.5 & 19.4 & 37.9 \\
    motion-CNN \cite{GkioxariM14} & 66.3 & 16.0 & 60.0 & 51.6 & 88.6 & 18.9 & 10.8 & 23.9 & 83.4 & 96.7 & 18.2 & 17.2 & 14.0 & 84.4 & 19.3 & 72.6 & 61.8 & 76.8 & 17.3 & 46.7 & 14.3 & 45.7 \\
    full \cite{GkioxariM14} & \textbf{79.1} & 33.4 & 53.9 & \textbf{60.3} & \textbf{99.3} & 18.4 & 26.2 & 42.0 & \textbf{92.8} & \textbf{98.1} & 29.6 & 24.6 & 13.7 & \textbf{92.9} & 42.3 & \textbf{67.2} & 57.6 & 66.5 & 27.9 & 58.9 & \textbf{35.8} & 53.3 \\
    \hline
    \hline
     our s-net & 66.2 & 45.7 & 54.6 & 42.2 & 83.9 & 4.2 & 33.5 & 31.7 & 75.0 & 76.6 & 24.8& 18.5 & 28.3 & 82.3 & 70.8 & 18.2 & 32.6 & 31.7 & 31.7 & 23.9 & 18.8 & 42.6 \\
     our t-net & 64.2 & 38.1 & 80.1 & 39.0 & 91.8 & 34.7 & 57.4 & 74.6 & 74.5 & 77.6 & 31.3 & 40.9 & 18.5 & 89.4 & 59.0 & 32.3 & 69.3 & 82.9 & 25.8 & 46.1 & 22.2 & 54.8 \\
     our full net & 76.4 & \textbf{49.7} & \textbf{80.3} & 43.0 & 92.5 & \textbf{24.2} & \textbf{57.7} & \textbf{70.5} & 78.7 & 77.2 & \textbf{31.7} & \textbf{35.7} & \textbf{27.0} & 88.8 & \textbf{76.9} & 29.8 & \textbf{68.6} & \textbf{72.8} & \textbf{31.5} & \textbf{44.4} & 26.2 & \textbf{56.4} \\
    \hline
  \end{tabular}
  }
  \vspace{1mm}
  \caption{Action detection results on the JHMDB dataset. We report \emph{frame-AP} and \emph{video-AP} for the spatial net (our s-net) and temporal net (our t-net), and their combination (our full net). We compare our method with the state-of-the-art performance \cite{GkioxariM14}  on this dataset.}
  \label{tbl:action-detection}
  \vspace{-4mm}
\end{table*}

\subsection{Evaluation on actionness estimation}

\textbf{Evaluation protocol.} We first evaluate the performance of our method on actionness estimation. Following \cite{ChenXXC14}, we select the mean average precision (mAP) to evaluate our approach. First, we plot $16 \times 16$ grids for images or $16 \times 16 \times 4$ grids for videos. Then, we score the patch or cuboid of each grid using the average of actionness confidence in this patch or cuboid. The patch or cuboid is treated as positive sample if its intersection over union (IoU) with respect to the ground truth bounding box is larger than 0.5 threshold. Finally, based on the scores and labels of patches or cuboids, we plot precision-recall (PR) curve and report average precision (AP) as the area under this curve for each test sample. mAP is obtained by taking average over all the test samples.

\textbf{Results.} We conduct experiments on images (Stanford40) and videos (UCF Sports and JHMDB). We first study the effect of multi-scale pyramid representation of image on actionness estimation and the results are reported in Figure \ref{fig:multiscale}. From these results, we see that the actionness maps of different scales are complementary to each other and the combination of them is useful for improving performance. We also report the computational time of different scales on the right of Figure \ref{fig:multiscale}. Thanks to the CUDA implementation of Caffe toolbox \cite{JiaSDKLGGD14}, it is efficient and only requires about $30ms$ to process an image with multi-scale pyramid representations using Tesla K40 GPU.

Table \ref{tbl:actionness-evaluation} shows the quantitative results of our method and the comparison with other approaches on three datasets. We only use A-FCN on the Stanford40 dataset as there is no motion information available in images. We separately investigate both M-FCN and A-FCN in videos, which are found complementary to each other. We first compare our method with previous actionness estimation method (L-CORF) \cite{ChenXXC14}. Our H-FCN outperforms L-CORF by around $7\%$ to $20\%$ on all these datasets, which indicates the effectiveness of fully convolutional networks. DPM \cite{FelzenszwalbGMR10} is another important baseline for both images and videos. It obtains the best performance on the dataset of Stanford40, which implies agent detection is important for actionness estimation. However, the performance of DPM on video datasets is much lower than that of H-FCN. This result may be ascribed to the fact that the human pose variations in image dataset is much smaller than in video datasets. Besides, the DPM lacks considering motion information.

\subsection{Evaluation on action proposal generation}

\textbf{Evaluation protocol.} Having evaluated the performance of H-FCNs on actionness estimation, we now apply actionness maps to produce action proposals. In the current implementation, we generate action proposals for each frame independently and therefore we conduct evaluation in frame level. There have been several works on action proposal generation \cite{OneataRVS14,JainGJBS14,YuY15}, but there is no standard evaluation protocol to evaluate these different proposal generation algorithms. We follow a recent comprehensive study on object proposals \cite{HosangBS14} and use proposal recall to measure the performance of action proposal methods. There are two kinds of measurements: (i) recall-number of proposal curve, which measures the detection rate versus the number of windows, with fixed IoU overlap threshold; (ii) recall-IoU overlap curve, which reports the detection rate versus IoU overlap, with fixed number of proposals.

\textbf{Results.} We conduct experiments on the datasets of Stanford40 and JHMDB, and the results are shown in Figure \ref{fig:actionproposal} and Figure \ref{fig:example-actionness}. From these results, we see that our estimated actionness maps are very effective for producing action proposals. We only need to generate $10$ boxes for each image on the dataset of Stanford40, and $4$ boxes for each frame on the dataset of JHMDB, to obtain 0.9 recall at IoU above 0.5. For higher IoU threshold (0.7), our method still achieves $0.5$-$0.6$ detection rate when producing 10 boxes for each image. We also separately report the performance of producing action proposals with the estimated maps by A-FCN and M-FCN on the dataset of JHMDB. We notice that A-FCN is better than M-FCN and the combination of them can further boost the performance. 

Next, we compare our method on action proposal generation with actionness estimation algorithm (L-CORF) \cite{ChenXXC14}, DPM \cite{FelzenszwalbGMR10}, and objectness method \cite{AlexeDF12}. These three methods use the same NMS score sampling to produce bounding boxes and only differ in how to generate the confidence maps for sampling. From the results in Figure \ref{fig:actionproposal}, we see that our method achieves comparable performance on images but much better performance on videos. Finally, we also compare our method with a recent action proposal method, namely STODP \cite{OneataRVS14}, on videos and our method outperforms this approach by a large margin. We also show several examples of actionness maps and action proposals in Figure \ref{fig:example-actionness}.

\subsection{Evaluation on action detection}

\textbf{Evaluation protocol.} Finally, we evaluate the performance of action detection using our generated action proposals. Following a recent work on action detection \cite{GkioxariM14}, we choose two evaluation criteria: \textbf{frame-AP} and \textbf{video-AP}. Frame-AP measures the area under the precision-recall curve of the detection for each frame. A detection is correct if the IoU with ground truth at that frame is greater than $0.5$ and the predicted label is correct. Video-AP measures the area under the precision-recall of the action tubes detection. A tube is correct if the mean of per-frame IoU value across the whole video is larger than $0.5$ and the action label is correctly predicted.

\textbf{Results.} We use the generated action proposals of each frame in previous subsections and perform action classification on these proposals. We choose the two-stream convolutional networks \cite{SimonyanZ14} as action classifiers due to their good performance on action recognition. As we generate action proposals for each frame independently, we first report the performance using frame-AP measurement and results are shown in Table \ref{tbl:action-detection}. We notice that temporal nets (t-net) outperform the spatial nets (s-net) on action detection, which is consistent with fact that temporal nets are better than spatial nets for action recognition \cite{SimonyanZ14}. Next, we generate action tubes for the whole video and report the performance evaluated by video-AP. To generate action tubes, we resort to the same temporal linking method in \cite{GkioxariM14}. The linking algorithm jointly considers the overlaps between detected regions of consecutive frames and their detection scores, and seeks a maximum temporal path over the video. The performance regarding action tubes are shown in Table \ref{tbl:action-detection} and there is a significant improvement (around 15\%) over frame based detection, which implies that the temporal structure is of great importance for action detection in videos. Finally, we compare our method with the state-of-the-art approach \cite{GkioxariM14} and our performance is better than theirs by about $3\%$ for both frame-AP and video-AP evaluation.


\section{Conclusions}
\label{sec:conclusion}

In this paper we have proposed a new deep architecture for efficient actionness estimation, called \emph{hybrid fully convolutional networks} (H-FCN). H-FCN is composed of appearance FCN (A-FCN) and motion FCN (M-FCN), which incorporates the static and dynamic visual cues for estimating actionness, respectively. Our method obtained the state-of-the-art performance for actionness estimation on three challenging datasets. In addition, we applied our estimated actionness maps on action proposal generation and action detection, which further demonstrates the effectiveness of estimated actionness maps on relevant video analysis tasks.

\section*{Acknowledgement}
Yu Qiao is the corresponding author and partially supported by Guangdong Innovative Research Program (2015B010129013, 2014B050505017) and Shenzhen Research Program (KQCX2015033117354153, JSGG20150925164740726, CXZZ20150930104115529). In addition, we gratefully acknowledge support through the ERC Advanced Grant VarCity.

\newpage
{\small
\bibliographystyle{ieee}
\bibliography{reference}
}

\end{document}